﻿\ifdefined\pdfminorversion
\fi
\ifdefined\pdfobjcompresslevel
\fi
\ifdefined\pdfcompresslevel
\fi
\documentclass{article}

\usepackage[final]{corl_2026} % CoRL 2026 camera-ready version.
\makeatletter
\renewcommand{\@conferencelocation}{Austin, Texas, USA}
\makeatother

\usepackage[utf8]{inputenc}
\DeclareUnicodeCharacter{03C0}{\ensuremath{\pi}} % Preserve the source-exported BibTeX title.
\usepackage[T1]{fontenc}
\usepackage{booktabs}
\usepackage{amsmath}
\usepackage{amsfonts}
\usepackage{nicefrac}
\usepackage{microtype}
\usepackage[table]{xcolor}
\definecolor{mydarkblue}{rgb}{0,0.08,0.45}
\hypersetup{
  pdftitle={LaWAM: Latent World Action Models for Efficient Dynamics-Aware Robot Policies},
  pdfauthor={Jialei Chen, Kai Wang, Kang Chen, Shuaihang Chen, Feng Gao, Wenhao Tang, Zhiyuan Li, Weilin Liu, Zhuyu Yao, Boxun Li, Yuanbo Xu, Chao Yu},
  linkcolor=mydarkblue,
  citecolor=mydarkblue,
  filecolor=mydarkblue,
  urlcolor=mydarkblue
}
\usepackage{float}
\usepackage{graphicx}
\usepackage{tikz}
\usepackage{subcaption}
\usepackage{wrapfig}
\usepackage{tabularx}
\usepackage{makecell}
\newcommand{\na}{\textemdash}
\usepackage{amssymb}
\title{LaWAM: Latent World Action Models for Efficient Dynamics-Aware Robot Policies}

\author{
\begin{tabular}[t]{c}
\normalfont\small
Jialei Chen{\textsuperscript{\textnormal{$\blacklozenge$,$\blacktriangledown$,$\bigstar$}}} \quad
Kai Wang{\textsuperscript{\textnormal{$\blacktriangleright$,$\blacktriangledown$}}} \quad
Kang Chen{\textsuperscript{\textnormal{$\clubsuit$,$\blacktriangledown$}}} \quad
Shuaihang Chen{\textsuperscript{\textnormal{$\blacksquare$,$\blacktriangledown$}}} \\
\normalfont\small
Feng Gao{\textsuperscript{\textnormal{$\spadesuit$,$\bigstar$}}} \quad
Wenhao Tang{\textsuperscript{\textnormal{$\spadesuit$}}} \quad
Zhiyuan Li{\textsuperscript{\textnormal{$\Diamond$}}} \quad
Weilin Liu{\textsuperscript{\textnormal{$\Diamond$}}} \\
\normalfont\small
Zhuyu Yao{\textsuperscript{\textnormal{$\Diamond$}}} \quad
Boxun Li{\textsuperscript{\textnormal{$\Diamond$}}} \quad
Yuanbo Xu{\textsuperscript{\textnormal{$\blacklozenge\dagger$}}} \quad
Chao Yu{\textsuperscript{\textnormal{$\spadesuit\dagger$}}} \\
\normalfont\footnotesize\itshape
\textsuperscript{$\spadesuit$}Tsinghua University \quad
\textsuperscript{$\blacklozenge$}Jilin University \quad
\textsuperscript{$\blacktriangleright$}Nankai University \quad
\textsuperscript{$\clubsuit$}Peking University \\
\normalfont\footnotesize\itshape
\textsuperscript{$\blacksquare$}Harbin Institute of Technology \quad
\textsuperscript{$\blacktriangledown$}Zhongguancun Academy \quad
\textsuperscript{$\bigstar$}Striding.AI \quad
\textsuperscript{$\Diamond$}Infinigence AI
\end{tabular}
}

\begin{document}
\maketitle

%===============================================================================
% \begin{abstract}
% 	Vision-Language-Action models (VLAs) ground robot policies in semantic understanding but often lack physical foresight into how actions reshape scenes.
% 	Recent World-Action Models (WAMs) introduce such foresight by generating visual futures, yet pixel-level prediction is costly and dominated by visual redundancy.
% 	We propose \textbf{\textit{LaWAM}}, a \textbf{La}tent \textbf{W}orld \textbf{A}ction \textbf{M}odel that moves future prediction from pixel reconstruction to a compact latent world-model interface.
% 	Derived from the latent action model paradigm, LaWAM repurposes the LAM forward decoder as latent world model, which expands policy-inferred latent actions into embodiment-grounded latent visual subgoals for action-chunk generation.
% 	Across benchmark and real-world manipulation evaluations, LaWAM achieves state-of-the-art or highly competitive performance against strong VLA and WAM baselines while maintaining low inference latency.
% 	LaWAM runs at 187\,ms per action-chunk prediction and achieves up to $24\times$ wall-clock speedup over pixel-space WAMs.
% \end{abstract}

\begin{abstract}
	Vision-Language-Action models (VLAs) leverage large-scale vision-language pretraining for semantic robot control, but often lack explicit foresight into how robot actions change the scene.
	World-Action Models (WAMs) address this limitation by conditioning policies on predicted futures, yet existing approaches typically rely on computationally expensive video generation with substantial pixel-level redundancy.
	We present \textbf{\textit{LaWAM}}, a \textbf{La}tent \textbf{W}orld \textbf{A}ction \textbf{M}odel that exposes predictive dynamics to robot policies through compact latent visual subgoals instead of reconstructed future video.
	At the core of LaWAM is a latent-action-conditioned \textbf{La}tent \textbf{W}orld \textbf{M}odel (LaWM).
	We obtain LaWM by training a latent action model on features from a pretrained vision foundation model and repurposing its forward decoder to predict future observation features.
	LaWAM then conditions action generation on these predicted latent visual subgoals to enable dynamics-aware robot control.
	LaWAM achieves state-of-the-art or competitive success rates (SRs) across LIBERO (98.6\% SR), RoboTwin (91.22\% SR), and real-world manipulation tasks while retaining low-latency inference.
	LaWAM runs in 187\,ms per action-chunk prediction and achieves up to $24\times$ lower wall-clock latency than pixel-space WAMs.
	Project website: \url{https://rlinf.github.io/LaWAM/}.
\end{abstract}


\section{Introduction}
\begin{wrapfigure}{r}{0.48\linewidth}
	\vspace{-1.2em}
	\centering
	\includegraphics[
		width=\linewidth,
		trim=0 0 0 6mm,
		clip
	]{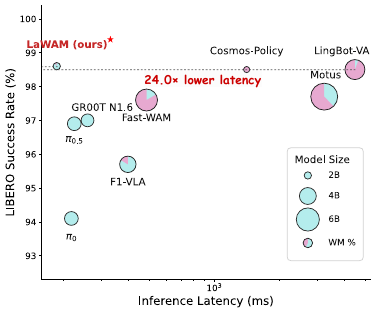}
	\vspace{-1.3em}
	\captionsetup{font=small}
	\caption{\textbf{Latency--success trade-off on LIBERO.} Latency for 10 denoising steps on an A100 GPU versus LIBERO success rate. The marker area denotes model size; the pink sector denotes world-modeling parameters.}
	% LaWAM matches the strongest LIBERO success rate while substantially reducing latency compared with pixel-space WAM baselines.
	\label{fig:libero_latency}
	\vspace{-0.8em}
\end{wrapfigure}

Vision-Language-Action models (VLAs) \cite{kim2024openvlaopensourcevisionlanguageactionmodel, kimOpenVLAOFTFinetuningVisionlanguageaction2025,https://doi.org/10.48550/arxiv.2410.24164, geminiroboticsteam2025geminiroboticsbringingai, gr00tn1_2025} have recently shown strong performance on robotic manipulation by transferring large-scale vision-language pretraining into action generation.
However, most current VLAs predict actions primarily from the current visual-language context, without explicitly modeling how the scene evolves under candidate actions \cite{mantisVersatileVisionlanguageaction2025, dialDecouplingIntentAction2026}.

\emph{World-Action Models} (WAMs) \cite{mimicvideoVideoactionModels2025, kim2026cosmospolicyfinetuningvideo, bi2025motusunifiedlatentaction, ye2026gigaworldpolicyefficientactioncenteredworldaction,li2026causalworldmodelingrobot, ye2026worldactionmodelszeroshot, lv2025f1visionlanguageactionmodelbridging} offer a natural way to introduce temporal dynamics by augmenting policies with predicted future observations or states as additional context.
However, current WAMs remain inefficient for manipulation policies.
First, many methods predict future images or videos, allocating substantial modeling capacity to pixel-level synthesis rather than compact action-relevant dynamics.
Second, iterative future generation introduces considerable inference latency; under the same evaluation setup used in Fig.~\ref{fig:libero_latency}, LingBot-VA \cite{li2026causalworldmodelingrobot} requires 4482\,ms for a single policy inference, whereas the representative VLA $\pi_{0.5}$ \cite{intelligence2025pi05visionlanguageactionmodelopenworld} requires only 220\,ms.
Third, effective future prediction for manipulation should expose the state change relevant to the next action chunk, rather than merely generating visually plausible task-consistent futures.

To address these limitations, we propose a \textbf{Latent World Model (LaWM)} for WAMs that predicts compact latent visual subgoals directly, rather than synthesizing future images or videos.
LaWM is a latent-action-conditioned dynamics model that predicts future observation features corresponding to the scene change required for the next action chunk.
By operating entirely in the latent space, LaWM models action-relevant dynamics efficiently without expensive or iterative pixel generation.

\begin{figure}[t]
	\centering
	\vspace{-0.8em}
	\includegraphics[
		width=\linewidth,
		trim=5 0 5 0mm,
		clip
	]{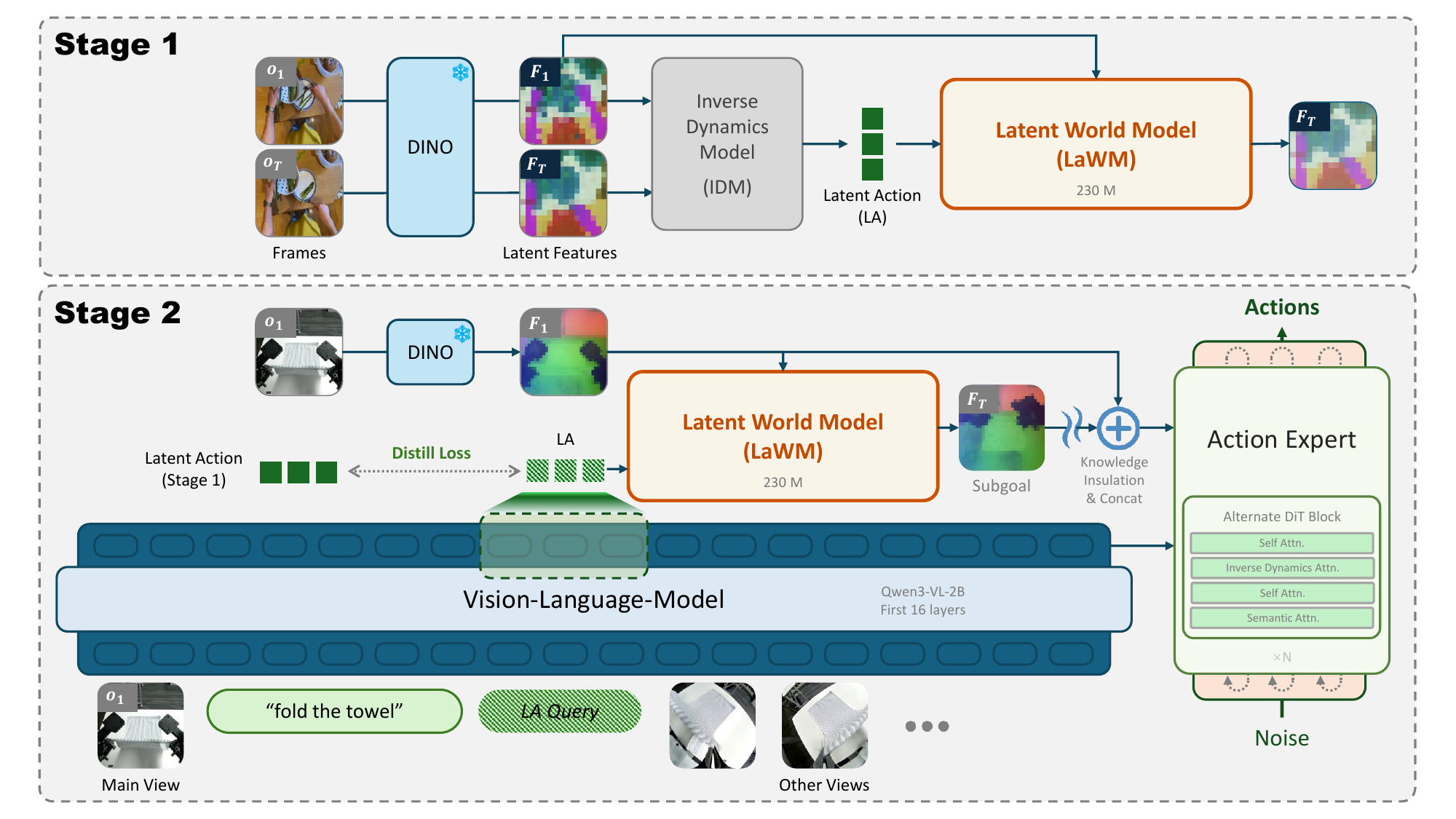}
	\vspace{-1.3em}
	\caption{\textbf{Overview of LaWAM.} Stage 1 learns a latent-action-conditioned world model from visual transitions: an inverse-dynamics encoder infers latent actions, and the decoder is retained as LaWM to predict future observation features. Stage 2 integrates LaWM into a VLA policy: latent-action distillation teaches the policy to drive LaWM, whose predicted latent visual subgoal is passed to an Alternate-DiT action expert for subgoal-conditioned action generation.}
	\label{fig:lawam}
	\vspace{-0.8em}
\end{figure}

We obtain LaWM by revisiting latent action models (LAMs) in the latent space of frozen visual encoders~\cite{bruce2024geniegenerativeinteractiveenvironments}, such as DINOv3~\cite{simeoni2025dinov3}.
A LAM contains an inverse-dynamics encoder that infers latent actions from visual transitions and a decoder that predicts the future latent state conditioned on the current latent state and latent action.
Prior latent-action-based VLAs mainly use this framework to learn embodiment-agnostic action representations, treating the decoder only as an auxiliary training component and discarding it after pretraining \cite{schmidtLearningActActions2024, bruce2024geniegenerativeinteractiveenvironments, ye2025latentactionpretrainingvideos, yang2025learning}.
In contrast, we repurpose the decoder as the core predictive module of LaWM, using latent actions to generate future observation features that serve as latent visual subgoals for action generation.

Building on this latent visual subgoal interface, we introduce the \textbf{Latent World Action Model (LaWAM)}, which conditions action generation on LaWM-predicted future dynamics. Fig.~\ref{fig:lawam} provides an overview of the full two-stage pretraining pipeline.
We first train LaWM as a latent-action-conditioned world model, then pretrain LaWAM with latent-action distillation for subgoal-conditioned action generation.
Overall, the training pipeline uses roughly 3,000 hours of robot videos and 1,500 hours of egocentric human videos.
At test time, the policy predicts a latent action, LaWM decodes it into a latent visual subgoal in a single forward pass, and an action expert generates the action chunk conditioned on both the current context and predicted subgoal.

After benchmark-specific post-training, LaWAM achieves state-of-the-art or competitive success against strong VLA and WAM baselines on LIBERO~\cite{liu2023libero}, RoboTwin~\cite{chen2025robotwin}, and real-world manipulation, including long-horizon tasks and deformable-object manipulation, with 2.3B parameters. 
Beyond accuracy, Fig.~\ref{fig:libero_latency} shows that LaWAM combines non-iterative latent prediction with a compact 230M-parameter LaWM, using about 95\% fewer world-modeling parameters than the 5B WAN backbone~\cite{wan2025wanopenadvancedlargescale}.
LaWAM runs in 187\,ms per action-chunk prediction and achieves up to $24\times$ lower wall-clock latency than pixel-space WAMs.

\section{Related Work}

\paragraph{Vision-Language-Action Models.}
VLAs transfer large-scale vision-language pretraining into robot control, giving policies strong semantic grounding over instructions, objects, and task compositions \cite{kim2024openvlaopensourcevisionlanguageactionmodel, kimOpenVLAOFTFinetuningVisionlanguageaction2025, geminiroboticsteam2025geminiroboticsbringingai, gr00tn1_2025, lv2025f1visionlanguageactionmodelbridging, mantisVersatileVisionlanguageaction2025}.
This semantic prior helps specify \emph{what} the robot should achieve, but it does not by itself provide explicit reasoning about \emph{how} the scene evolves under embodied interaction.
Several recent VLAs adopt query-based designs to incorporate predictive information through learned queries and feature-alignment objectives \cite{zhengFLARERobotLearning2025, gr00tn1_2025, beingbeyond2026beingh07, su2026worldguidanceworldmodeling}.
For example, Mantis and VLA-JEPA use future prediction as auxiliary supervision~\cite{mantisVersatileVisionlanguageaction2025,vlajepaEnhancingVisionlanguageaction2026}.
LaWAM instead retains the forward decoder at inference and explicitly conditions action generation on its predicted spatial future; Appendix~\ref{app:predictive_interfaces} details this distinction.

\paragraph{World-Action Models.}
World-Action Models (WAMs) jointly model future video and actions conditioned on observations and task instructions \cite{mimicvideoVideoactionModels2025, bi2025motusunifiedlatentaction, ye2026worldactionmodelszeroshot, kim2026cosmospolicyfinetuningvideo, li2026causalworldmodelingrobot, ye2026gigaworldpolicyefficientactioncenteredworldaction, fastwamNeedTesttimeFutureImagination2026}.
Pixel-space WAMs provide explicit physical foresight through generated images or videos, but they inherit costly iterative generation and substantial appearance-level redundancy \cite{kim2026cosmospolicyfinetuningvideo, li2026causalworldmodelingrobot, jiang2026wovr}.
$\pi_{0.7}$ similarly conditions robot policies on visual subgoals, but these subgoals are produced by a separate iterative pixel-space model \cite{physicalIntelligencePi07Steerable2026}.
Efficiency-oriented approaches such as Fast-WAM and Giga-World Policy reduce generation cost by making future prediction auxiliary or action-centered, which weakens the role of predicted dynamics as an explicit conditioning signal during test-time action generation~\cite{fastwamNeedTesttimeFutureImagination2026,ye2026gigaworldpolicyefficientactioncenteredworldaction}.
Concurrent analyses likewise suggest that pretrained visual latent spaces provide stronger policy-facing dynamics representations than pixel-reconstruction spaces~\cite{nilakshReconstructionSemantics2026}.
While LDA-1B also models dynamics in a structured DINO latent space, it jointly denoises future visual states and action chunks through a diffusion-style process \cite{lyu2026lda1bscalinglatentdynamics}.
LaWAM jointly models future visual features and actions, representing the future by the endpoint of each fixed-physical-horizon action chunk. It decodes this latent visual subgoal in a single forward pass and directly conditions action generation on it at inference.

\paragraph{Latent Action Models.}
Latent action models learn compact transition variables from unlabeled videos through latent inverse dynamics and forward prediction \cite{schmidtLearningActActions2024, bruce2024geniegenerativeinteractiveenvironments, ye2025latentactionpretrainingvideos, yang2025learning}.
Recent VLA and world-modeling methods use such variables as cross-embodiment action representations for policy learning and future prediction \cite{ye2025latentactionpretrainingvideos, bu2025univlalearningacttaskcentric, gaoAdaWorldLearningAdaptable2025, gao2026dreamdojogeneralistrobotworld}.
Most prior work focuses primarily on the latent action space itself---its discreteness or continuity, regularization, and downstream transfer.
A complementary opportunity lies in the decoder: as concurrently observed by \citet{learningLatentActionWorldModelsWild2026}, the decoder of a latent action model already implements a latent action-conditioned world model.
LaWAM builds on this observation by repurposing the decoder as a policy-facing latent dynamics interface, where latent actions are expanded into embodiment-grounded future observation features that directly condition action generation.

\section{Method}
\label{sec:method}

\subsection{Problem Formulation}
Let $o$ denote the current observation, $l$ the task instruction, and $a_{1:T}$ an action chunk over a fixed physical horizon $\tau$. A standard VLA directly models $p(a_{1:T}\mid o,l)$. A WAM jointly models future video and actions, $p(o_{1:T},a_{1:T}\mid o,l)$ \cite{ye2026worldactionmodelszeroshot}. For chunk-level control, we consider the horizon observation $o_T$ and factor the corresponding joint distribution as
\begin{equation}
	\underbrace{p(a_{1:T}, o_T\mid o, l)}_{\text{Joint}}
	=
	\underbrace{p(o_T\mid o, l)}_{\text{Future Prediction}}
	\underbrace{p(a_{1:T}\mid o,l,o_T)}_{\text{Future-Conditioned Actions}}.
\end{equation}
The second term plays an inverse-dynamics role, mapping the current observation and a desired visual future to actions while retaining the task instruction.

LaWAM realizes this joint modeling in a frozen visual feature space. Let $f_{\psi}$ be the frozen encoder, and define $u=f_{\psi}(o)$ and $u_T=f_{\psi}(o_T)$. Our joint target is $(u_T,a_{1:T})$: the spatial visual features at the chunk endpoint and the actions over the same physical interval. We model this temporally coarse latent future rather than reconstructing every intervening video frame. To construct the future predictor, we first learn a latent action model over feature pairs:
\begin{equation}
	\label{eq:lam_interface}
	z\sim q_{\phi}(z\mid u,u_T),
	\qquad
	\tilde{u}_T=\mathrm{LaWM}_{\omega}(u,z).
\end{equation}
The distribution $q_{\phi}$ is the latent-action posterior: it acts as a latent inverse-dynamics model, inferring $z$ from the observed transition $(u,u_T)$. The decoder then predicts the horizon feature from the current feature and $z$, and we retain this decoder as the Latent World Model (LaWM).

At inference time, the ground-truth future feature $u_T$ is unavailable, so the policy predicts a latent action and decodes its visual consequences through LaWM. The resulting joint world--action model factors as
\begin{equation}
	\label{eq:lawam_factorization}
	\underbrace{p(a_{1:T},\hat{u}_T,\hat{z}\mid o,l)}_{\text{LaWAM}}
	=
	\underbrace{p_{\theta}(\hat{z}\mid o,l)}_{\text{Policy Prior}}
	\underbrace{p_{\omega}(\hat{u}_T\mid u,\hat{z})}_{\text{LaWM}}
	\underbrace{p_{\eta}(a_{1:T}\mid o,l,u,\hat{u}_T)}_{\text{Action Expert}}.
\end{equation}
Here $p_{\theta}$ is implemented as a regressor producing the point prediction $\hat{z}$, $p_{\omega}$ deterministically decodes it into $\hat{u}_T=\mathrm{LaWM}_{\omega}(u,\hat{z})$, and $p_{\eta}$ generates the action chunk. The first two conditional distributions therefore represent deterministic mappings. We use $u_T$ for the training target and $\hat{u}_T$ for the policy-driven subgoal.

The action expert in Eq.~\ref{eq:lawam_factorization} conditions on both the current features $u$ and the predicted endpoint $\hat{u}_T$, connecting latent future prediction to action generation.

\subsection{Latent World Model}
The first stage learns LaWM as the forward decoder of a latent action model. Each training example contains a current observation $o$ and a horizon observation $o_T$ sampled after the physical interval $\tau$. After encoding them as $(u,u_T)$, the inverse-dynamics encoder infers a latent action $z\sim q_{\phi}(z\mid u,u_T)$. The decoder then uses $(u,z)$ to predict the horizon feature, $\tilde{u}_T=\mathrm{LaWM}_{\omega}(u,z)$.

This training setup gives LaWM a simple role: given the current latent state and a latent action, predict the latent future state. The target $u_T$ directly supervises the decoder, and the inferred latent action $z$ serves as the teacher signal for the stage-two policy prior.

We add one auxiliary signal during this pretraining stage. A lightweight predictor $g$ maps the current end-effector state $s$ and latent action $z$ to the horizon state $s_T$, encouraging $z$ to encode embodied motion rather than only visual appearance change. After pretraining, we discard this auxiliary head and keep the decoder as LaWM; the posterior encoder is used only to produce teacher latent actions for policy training. LaWM architecture details are provided in Appendix~\ref{app:lawm_arch}.

We train the latent action model with a forward-prediction objective and KL regularization,
\begin{equation}
	\label{eq:lam_objective}
	\mathcal{L}_{\mathrm{LAM}}
	=
	\mathcal{L}_{\mathrm{wm}}^{(1)}
	+ \mathcal{L}_{\mathrm{aux}}
	+ \beta\,\mathbb{E}\!\left[D_{\mathrm{KL}}\!\left(q_{\phi}(z\mid u, u_T) \,\|\, \mathcal{N}(0, I)\right)\right],
\end{equation}
where $\mathcal{L}_{\mathrm{wm}}^{(1)}=\mathbb{E}[\lVert \tilde{u}_T-u_T\rVert_2^2]$ trains LaWM to match the horizon feature, and $\mathcal{L}_{\mathrm{aux}}=\mathbb{E}[\lVert g(s,z)-s_T\rVert_2^2]$ trains the auxiliary state predictor. Expectations are taken over training samples and, where applicable, posterior latent-action sampling. The superscript $(1)$ identifies Stage-1 future prediction driven by the posterior latent action $z$. The KL term regularizes the latent-action space so that it can later be modeled by the policy prior. Because both $u_T$ and $s_T$ are defined by the same physical interval $\tau$, the learned subgoal corresponds to a consistent amount of elapsed motion rather than a dataset-specific frame offset.

\begin{figure}[t]
	\centering
	\vspace{-0.8em}
	\includegraphics[width=0.95\linewidth]{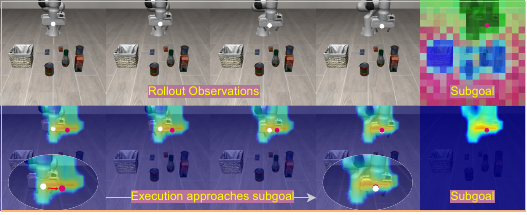}
	\caption{\textbf{Subgoal-guided chunk execution.}
		The top row shows observations within one executed LIBERO chunk together with the predicted latent subgoal; the bottom row overlays subgoal-derived robot-arm heatmaps, illustrating how the executed motion approaches the predicted subgoal region.}
	\vspace{-0.8em}
	\label{fig:libero_chunk}
\end{figure}

\subsection{Latent World Action Model}
\label{sec:lawam}
Stage 2 integrates the pretrained LaWM into a language-conditioned robot policy. Since the Stage-1 posterior requires the future observation, we learn a policy prior $p_{\theta}(\hat{z}\mid o,l)$ that predicts the latent action from the current observation and instruction. Learned VLM queries are aggregated and projected to $\hat{z}$, which LaWM decodes in a single forward pass into $\hat{u}_T=\mathrm{LaWM}_{\omega}(u,\hat{z})$. An Alternate-DiT action expert~\cite{gr00tn1_2025} alternates attention between VLM semantic context and the spatial dynamics context $(u,\hat{u}_T)$ to generate the action chunk. Fig.~\ref{fig:libero_chunk} illustrates the resulting subgoal-guided execution; architecture details are provided in Appendix~\ref{app:lawam_arch}.

\paragraph{Joint training objectives.}
The policy prior, LaWM, and action expert are jointly trained with
\begin{equation}
	\label{eq:lawam_objective}
	\mathcal{L}_{\mathrm{LaWAM}}
	=
	\lambda_{\mathrm{distill}}\mathcal{L}_{\mathrm{distill}}
	+ \lambda_{\mathrm{wm}}\mathcal{L}_{\mathrm{wm}}^{(2)}
	+ \mathcal{L}_{\mathrm{act}}.
\end{equation}
Here $\mathcal{L}_{\mathrm{distill}}=\mathbb{E}[\lVert\hat{z}-z\rVert_2^2]$ aligns the policy prediction with teacher latent actions from the frozen Stage-1 posterior, preserving the latent interface on which LaWM was pretrained. The future loss $\mathcal{L}_{\mathrm{wm}}^{(2)}=\mathbb{E}[\lVert\hat{u}_T-u_T\rVert_2^2]$ supervises LaWM under policy-predicted latent actions, while $\mathcal{L}_{\mathrm{act}}$ learns action generation through conditional flow matching. The action expert uses the predicted subgoal during both training and inference; the ground-truth future provides supervision only.

Knowledge Insulation (KI)~\cite{driess2025knowledgeinsulatingvisionlanguageactionmodels} passes $\operatorname{sg}[\hat{u}_T]$ to the action expert, blocking action-loss gradients through the subgoal branch while retaining predictive supervision. Initialization, gradient routing, and training schedules are detailed in Appendix~\ref{app:training_details}. For mixed-frequency data, physical-time encoding aligns native-rate action chunks with the subgoal horizon $\tau$ (Appendix~\ref{app:physical_time}).

\section{Experiments}

We use a modest pretraining setup built from open-source data \cite{hoqueEgoDexLearningDexterous2025,grauman2022ego4dworld3000hours,laiLEGOLearningEGOcentric2024,agibotworldcontributors2025agibotworldcolosseolargescale,wu2025robomind,RoboCOINReport,open_x_embodiment_rt_x_2023,khazatskyDROIDLargescaleInthewild2025}.
LaWM is pretrained on roughly 3,000 hours of robot videos and 1,500 hours of egocentric human videos, while LaWAM policy integration uses only robot trajectories with language instructions; human videos contribute only through the dynamics prior learned by LaWM.
We evaluate LaWAM along five axes: simulated benchmark performance, inference efficiency, real-world transfer, latent-dynamics behavior, and component contributions. Benchmark-specific protocols are detailed in Appendices~\ref{app:libero_protocol}, \ref{app:robotwin_protocol}, and \ref{app:real_world_protocol}, with visualization protocols in Appendix~\ref{app:visualization}.

\subsection{LIBERO}

We first evaluate on the four standard LIBERO suites \cite{liu2023libero}. As shown in Table~\ref{tab:comparison}, LaWAM achieves the best average success rate among the compared VLA, latent-action, and WAM baselines; the gap over latent-action baselines suggests that compact action tokens are more effective when expanded into spatially structured latent visual subgoals.

\paragraph{Inference efficiency.}
LaWAM predicts an action chunk in 187\,ms, compared with 486\,ms for Fast-WAM and 4,482\,ms for LingBot-VA (Table~\ref{tab:comparison}), achieving up to $24\times$ lower latency among the compared pixel-space WAMs (Fig.~\ref{fig:libero_latency}). LaWAM combines a compact 230M-parameter LaWM with single-pass latent subgoal prediction, avoiding iterative video generation while retaining future-conditioned action generation. The world model uses about 95\% fewer parameters than a 5B WAN backbone, and the full policy contains 2.3B parameters. Measurement protocols, parameter accounting, and training costs are provided in Appendix~\ref{app:efficiency_protocol}.

\begin{table*}[t]
  \centering
  \scriptsize
  \vspace{-0.8em}
  \caption{\textbf{LIBERO benchmark results over 50 trials per task.} Latency is model-only wall-clock time per action chunk, excluding simulator and robot execution overhead. The best result in each metric is shown in bold, the second-best is underlined, and large models or high latency are highlighted in red.}
  \label{tab:comparison}
  \definecolor{mainstreambg}{RGB}{237,241,251}
  \definecolor{latentbg}{RGB}{235,248,244}
  \definecolor{querybg}{RGB}{241,235,248}
  \definecolor{pixelbg}{RGB}{251,244,236}
  \definecolor{lawambg}{RGB}{236,244,249}
  \newcolumntype{Y}{>{\centering\arraybackslash}X}
  \newcommand{\rowstrut}{\rule{0pt}{2.3ex}}
  \newcommand{\bestval}[1]{\textbf{#1}}
  \newcommand{\secondval}[1]{\underline{#1}}
  \newcommand{\alertval}[1]{\textcolor{red!75!black}{#1}}
  \newcommand{\bestnum}[1]{\bestval{#1}}
  \newcommand{\secondnum}[1]{\secondval{#1}}
  \newcommand{\alertnum}[1]{\alertval{#1}}
  \newcommand{\bgrow}[9]{\rowstrut #2 & #3 & #4 & #5 & #6 & #7 & #8 & #9\\}
  \newcommand{\grouprow}[2]{\multicolumn{8}{>{\columncolor{#1}}c}{\rowstrut\textsc{#2}}\\}
  \setlength{\tabcolsep}{3pt}
  \renewcommand{\arraystretch}{1.0}
  \begin{tabularx}{\textwidth}{@{}>{\raggedright\arraybackslash}p{0.17\textwidth}>{\centering\arraybackslash}p{0.11\textwidth}>{\centering\arraybackslash}p{0.12\textwidth}YYYYY@{}}
    \toprule
    Method & Model Size & Latency (ms) & Long & Goal & Object & Spatial & Average \\
    \midrule
    \grouprow{mainstreambg}{Mainstream VLA}
    \bgrow{mainstreambg}{OpenVLA-OFT~\cite{kimOpenVLAOFTFinetuningVisionlanguageaction2025}}{\alertval{7B}}{\na}{94.5}{97.9}{98.4}{97.6}{97.1}
    \bgrow{mainstreambg}{$\pi_{0}$~\cite{https://doi.org/10.48550/arxiv.2410.24164}}{3.5B}{\secondnum{220}}{88.4}{94.4}{96.8}{98.0}{94.4}
    \bgrow{mainstreambg}{$\pi_{0.5}$~\cite{intelligence2025pi05visionlanguageactionmodelopenworld}}{3.5B}{\secondnum{220}}{92.4}{98.0}{98.2}{\secondnum{98.8}}{96.9}
    \bgrow{mainstreambg}{GR00T-N1.6~\cite{gr00tn1_2025}}{3.3B}{259}{94.4}{97.5}{98.5}{97.7}{97.0}
    \addlinespace[2pt]
    \grouprow{latentbg}{Latent Action Based methods}
    \bgrow{latentbg}{LAPA~\cite{ye2025latentactionpretrainingvideos}}{\alertval{7B}}{\na}{55.4}{58.8}{74.6}{73.8}{65.7}
    \bgrow{latentbg}{UniVLA~\cite{bu2025univlalearningacttaskcentric}}{\alertval{7B}}{\na}{92.0}{95.6}{96.8}{96.5}{95.2}
    \bgrow{latentbg}{Mantis~\cite{mantisVersatileVisionlanguageaction2025}}{\alertval{5.8B}}{\na}{94.2}{94.4}{99.2}{\secondnum{98.8}}{96.7}
    \bgrow{latentbg}{VLA-JEPA~\cite{vlajepaEnhancingVisionlanguageaction2026}}{3B}{\na}{95.8}{97.2}{99.6}{96.2}{97.2}
    \addlinespace[2pt]
    \grouprow{pixelbg}{WAMs in Pixel Space}
    \bgrow{pixelbg}{F1~\cite{lv2025f1visionlanguageactionmodelbridging}}{4B}{399.0}{91.3}{95.4}{97.8}{98.2}{95.7}
    \bgrow{pixelbg}{Motus~\cite{bi2025motusunifiedlatentaction}}{\alertval{8B}}{\alertnum{3231}}{\secondnum{97.6}}{96.6}{\secondnum{99.8}}{96.8}{97.7}
    \bgrow{pixelbg}{Cosmos-Policy~\cite{kim2026cosmospolicyfinetuningvideo}}{2.1B}{\alertnum{1413}}{\secondnum{97.6}}{\secondnum{98.2}}{\bestnum{100.0}}{98.1}{\secondnum{98.5}}
    \bgrow{pixelbg}{LingBot-VA~\cite{li2026causalworldmodelingrobot}}{\alertval{5.5B}}{\alertnum{4482}}{\bestnum{98.5}}{97.2}{99.6}{98.5}{\secondnum{98.5}}
    \bgrow{pixelbg}{Fast-WAM~\cite{fastwamNeedTesttimeFutureImagination2026}}{\alertval{6B}}{486}{95.2}{97.0}{\bestnum{100.0}}{98.2}{97.6}
    \addlinespace[2pt]
    \grouprow{lawambg}{Latent World Action Model}
    \bgrow{lawambg}{LaWAM}{2.3B}{\bestnum{187}}{97.0}{\bestnum{98.4}}{99.6}{\bestnum{99.4}}{\bestnum{98.6}}
    \bottomrule
  \end{tabularx}
\end{table*}

\begin{table*}[t]
    \centering
    \tiny
    % \vspace{-0.8em}
    \caption{\textbf{RoboTwin benchmark results over 100 trials per task.} Fast-WAM and LingBot-VA are re-evaluated from their open weights on H100 GPUs; the remaining baseline results are taken from Fast-WAM~\cite{fastwamNeedTesttimeFutureImagination2026} and GigaWorld-Policy~\cite{ye2026gigaworldpolicyefficientactioncenteredworldaction}. Best results are marked in bold.}
    \label{tab:robotwin}
    \setlength{\tabcolsep}{2pt}
    \resizebox{\textwidth}{!}{
        \begin{tabular}{lcc|cc|cc|cc|cc|cc}
            \toprule
            Task                  & \multicolumn{2}{c}{Fast-WAM} & \multicolumn{2}{c}{GigaWorld-Policy} & \multicolumn{2}{c}{LingBot-VA} & \multicolumn{2}{c}{$\pi_{0.5}$} & \multicolumn{2}{c}{Motus} & \multicolumn{2}{c}{LaWAM}                                                                 \\
            \cmidrule(lr){2-3}\cmidrule(lr){4-5}\cmidrule(lr){6-7}\cmidrule(lr){8-9}\cmidrule(lr){10-11}\cmidrule(lr){12-13}
                                  & Clean                        & Rand.                                & Clean                          & Rand.                           & Clean                     & Rand.                     & Clean & Rand. & Clean & Rand. & Clean          & Rand.        \\
            \midrule
            Move Can Pot          & 95                           & 92                                   & 76                             & 78                              & 93                        & \textbf{93}               & 51    & 55    & 34    & 74    & \textbf{98}    & \textbf{93}  \\
            Move Stapler Pad      & 84                           & 63                                   & 92                             & 82                              & 59                        & 71                        & 56    & 42    & 83    & 85    & \textbf{94}    & \textbf{87}  \\
            Open Laptop           & \textbf{100}                 & \textbf{100}                         & 96                             & 98                              & 96                        & 90                        & 90    & 96    & 95    & 91    & \textbf{100}   & \textbf{100} \\
            Pick Dual Bottles     & \textbf{100}                 & 94                                   & 86                             & 86                              & \textbf{100}              & \textbf{100}              & 93    & 63    & 96    & 90    & \textbf{100}   & 95           \\
            Place A2B Left        & 94                           & 89                                   & 94                             & 88                              & 96                        & \textbf{94}               & 87    & 82    & 88    & 79    & \textbf{98}    & 91           \\
            \addlinespace[2pt]
            \multicolumn{13}{c}{$\cdots$ (50 tasks in total)}                                                                                                                                                                                                                                      \\
            \addlinespace[2pt]
            Place Container Plate & 98                           & \textbf{100}                         & 98                             & 96                              & 97                        & 98                        & 99    & 95    & 98    & 99    & \textbf{100}   & \textbf{100} \\
            Place Dual Shoes      & 88                           & 88                                   & 96                             & 84                              & 94                        & 81                        & 75    & 75    & 93    & 87    & \textbf{98}    & \textbf{94}  \\
            Place Object Basket   & 82                           & 90                                   & 90                             & \textbf{92}                     & 89                        & 85                        & 80    & 76    & 81    & 87    & \textbf{92}    & 90           \\
            Place Object Scale    & 83                           & \textbf{88}                          & 88                             & 80                              & \textbf{98}               & 87                        & 86    & 80    & 88    & 85    & 95             & \textbf{88}  \\
            Place Shoe            & 97                           & 98                                   & 98                             & 96                              & 99                        & 97                        & 92    & 93    & 99    & 97    & \textbf{100}   & \textbf{100} \\
            Put Bottles Dustbin   & 93                           & 82                                   & 72                             & 70                              & 82                        & 87                        & 84    & 79    & 81    & 79    & \textbf{94}    & \textbf{92}  \\
            Put Object Cabinet    & \textbf{94}                  & \textbf{82}                          & 74                             & 74                              & 87                        & 80                        & 80    & 79    & 88    & 71    & 90             & \textbf{82}  \\
            Scan Object           & \textbf{96}                  & 86                                   & 60                             & 64                              & 90                        & 89                        & 72    & 65    & 67    & 66    & \textbf{96}    & \textbf{90}  \\
            Stack Bowls Two       & 90                           & 96                                   & 96                             & 92                              & 98                        & \textbf{99}               & 95    & 96    & 98    & 98    & \textbf{100}   & \textbf{99}  \\
            \midrule
            \textbf{Average}      & 91.98                        & 90.52                                & 86.36                          & 85.04                           & 91.50                     & \textbf{90.92}            & 82.74 & 76.76 & 88.66 & 87.02 & \textbf{92.64} & 89.80        \\
            \bottomrule
        \end{tabular}
    }
\end{table*}

\subsection{RoboTwin}
RoboTwin 2.0 \cite{chen2025robotwin} evaluates coordinated bimanual manipulation across 50 tasks, allowing us to test whether non-iterative latent world modeling scales beyond single-arm LIBERO tasks. Table~\ref{tab:robotwin} shows that LaWAM achieves the best clean-scene average and remains close to the strongest pixel-space WAMs under randomized scenes. These results suggest that latent subgoal prediction scales to complex bimanual manipulation while avoiding heavy test-time rollout cost of pixel-space WAMs.

\begin{figure}[t]
	\centering
	\vspace{-0.8em}
	\input{picture/real_world_rollouts_combined.tex}
	\caption{\textbf{Real-world manipulation on two robot platforms.} (a) Pick-and-Place and (b) Drawer Opening on Franka; (c) Towel Folding and (d) Clean Table on Quanta X1. Frames progress from left to right, with Clean Table continuing on the next row through eight object-level subtasks.}
	\label{fig:real_world_rollouts}
	\vspace{-0.8em}
\end{figure}

\subsection{Real-World Experiments}

\begin{table}[t]
    \centering
    \small
    \caption{\textbf{Real-world manipulation success rates (\%).} Trial counts are per method and task or condition. Light denotes additional red, green, or purple illumination at evaluation; each method uses the same checkpoint for a given task with and without Light. \textemdash\ denotes not evaluated. The 95\% Wilson confidence intervals for the 50-trial evaluations are provided in Appendix Table~\ref{tab:real_world_confidence}.}
    \label{tab:real_world_success}
    \setlength{\tabcolsep}{4pt}
    \begin{tabular*}{\linewidth}{@{\extracolsep{\fill}}lcccccc@{}}
        \toprule
        & \multicolumn{3}{c}{30 trials per task} & \multicolumn{3}{c}{50 trials per condition} \\
        \cmidrule(lr){2-4}\cmidrule(lr){5-7}
        Method & \makecell{Pick-and-\\Place} & \makecell{Open\\Drawer} & \makecell{Fold\\Towel} & \makecell{Fold Towel\\--Light} & \makecell{Clean\\Table} & \makecell{Clean Table\\--Light} \\
        \midrule
        $\pi_{0.5}$ & 86.7 & 80.0 & 83.3 & 46 & 28 & 10 \\
        GR00T-N1.6 & 83.3 & 76.7 & 46.7 & \textemdash & \textemdash & \textemdash \\
        Fast-WAM & 56.7 & 63.3 & 70.0 & 14 & 6 & 0 \\
        LingBot-VA & 76.7 & 83.3 & 0.0 & \textemdash & \textemdash & \textemdash \\
        LaWAM & \textbf{93.3} & \textbf{86.7} & \textbf{90.0} & \textbf{68} & \textbf{42} & \textbf{24} \\
        \bottomrule
    \end{tabular*}
\end{table}

We evaluate real-world manipulation on two robot platforms using four tasks: Pick-and-Place, Drawer Opening, Towel Folding, and Clean Table. These tasks span rigid and deformable objects, with Clean Table requiring completion of an eight-subtask sequence. To assess robustness to lighting changes, we additionally evaluate Towel Folding and Clean Table under red, green, or purple illumination. For each method and task, both lighting conditions use the same checkpoint trained with color augmentation. Fig.~\ref{fig:real_world_rollouts} illustrates the four tasks.

As shown in Table~\ref{tab:real_world_success}, LaWAM achieves the highest SR among the evaluated methods in all six conditions. On Clean Table, it achieves 42\% SR, compared with 28\% for $\pi_{0.5}$ and 6\% for Fast-WAM. Under lighting perturbations, it retains 68\% SR on Fold Towel--Light and 24\% on Clean Table--Light. These results extend LaWAM's strong performance on the three initial tasks to multi-stage manipulation and altered lighting. Training details, evaluation protocols, and confidence intervals are provided in Appendix~\ref{app:real_world_protocol}.

\begin{figure}[t]
	\centering
	\vspace{-0.3em}
	\includegraphics[width=\linewidth]{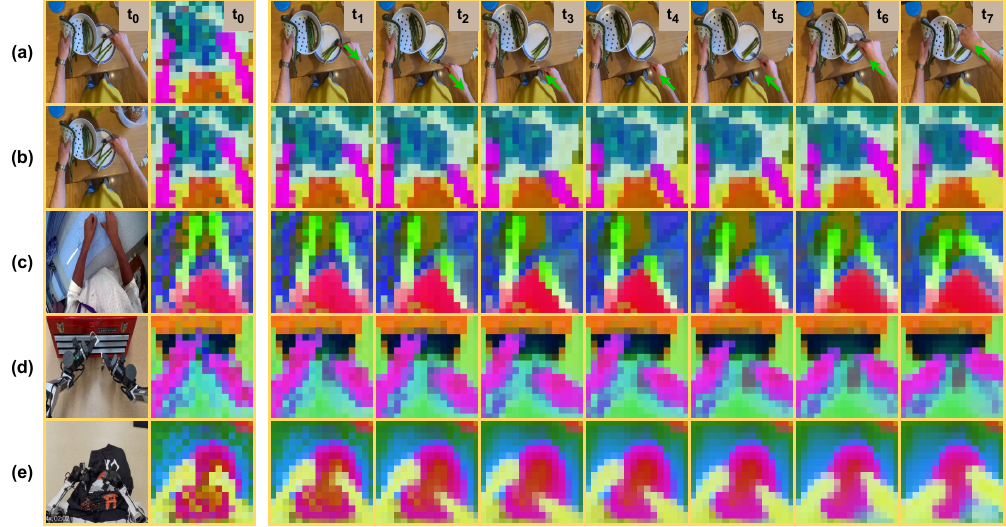}
	\caption{\textbf{Cross-embodiment open-loop LaWM rollouts from shared latent actions.} (a) We infer a latent-action trajectory from the source video. (b)--(e) Given only one initial observation in each environment or embodiment, LaWM applies the same extracted latent actions to generate context-specific latent rollouts. Panels (d) and (e) use unseen screenshots from \emph{pi.website}.}
	\label{fig:lam_intro_pca}
	\vspace{-0.8em}
\end{figure}

\subsection{LaWM Dynamics Analysis}

We next analyze whether LaWM captures coherent dynamics rather than merely providing an auxiliary feature for the policy. Open-loop analyses in Fig.~\ref{fig:lam_intro_pca} show that applying the same latent action to unseen environments and embodiments produces coherent latent-space changes. This cross-embodiment rollout suggests two complementary properties: the latent action captures an embodiment-agnostic visual transition, while LaWM grounds it in the current latent visual state, which retains the embodiment-specific information needed to realize the dynamics. This also clarifies why using latent actions alone as the final policy interface can be limiting: the latent action becomes most useful after LaWM expands it into a visual subgoal grounded in the current embodiment. Appendix~\ref{app:lawm_analysis} provides additional open-loop rollout and cross-embodiment visualizations of LaWM's dynamics-modeling behavior.

\subsection{Component Ablations}
We evaluate component contributions on LIBERO-10. Full LaWAM achieves 97.0\% SR, while LaWAM without Stage-2 pretraining achieves 96.4\%. We use the latter as the reference for the remaining component ablations, all of which omit Stage-2 pretraining. Table~\ref{tab:component_ablations} separates the training mechanisms from variants of the predictive interface; each row specifies an independent configuration.

\begin{table}[t]
	\centering
	\footnotesize
	\setlength{\tabcolsep}{3pt}
	\renewcommand{\arraystretch}{1.12}
	\caption{\textbf{Component ablations on LIBERO-10.} Only Full LaWAM includes Stage-2 pretraining; all other rows omit it. LaWM indicates whether the forward decoder is retained during training; future prediction denotes future-feature supervision, and subgoal conditioning indicates whether the predicted subgoal conditions action generation. LA denotes latent action. ``Only'' in variant names refers to auxiliary supervision; all variants retain the action loss. $\checkmark$: enabled; $\times$: removed; \textemdash: not applicable. KI blocks action-loss gradients through the predicted subgoal.}
	\label{tab:component_ablations}
	\begin{tabular*}{\linewidth}{@{\extracolsep{\fill}}lcccccc@{}}
		\toprule
		Setting & LaWM & \makecell{Future\\prediction} & \makecell{LA\\distillation} & \makecell{Subgoal\\conditioning} & KI & SR ($\%$) $\uparrow$ \\
		\midrule
		\textbf{Full LaWAM} & $\checkmark$ & $\checkmark$ & $\checkmark$ & $\checkmark$ & $\checkmark$ & \textbf{97.0} \\
		\midrule
		\makecell[l]{w/o Stage-2 Pretraining} & $\checkmark$ & $\checkmark$ & $\checkmark$ & $\checkmark$ & $\checkmark$ & 96.4 \\
		w/o KI & $\checkmark$ & $\checkmark$ & $\checkmark$ & $\checkmark$ & $\times$ & 94.2 \\
		w/o LA Distillation & $\checkmark$ & $\checkmark$ & $\times$ & $\checkmark$ & $\checkmark$ & 93.4 \\
		w/o KI \& LA Distillation & $\checkmark$ & $\checkmark$ & $\times$ & $\checkmark$ & $\times$ & 93.2 \\
		w/o Subgoal Conditioning & $\checkmark$ & $\checkmark$ & $\checkmark$ & $\times$ & \textemdash & 94.6 \\
		Future Prediction & $\checkmark$ & $\checkmark$ & $\times$ & $\times$ & \textemdash & 93.4 \\
		LA Distillation Only & $\times$ & $\times$ & $\checkmark$ & $\times$ & \textemdash & 95.2 \\
		Base VLA & $\times$ & $\times$ & $\times$ & $\times$ & \textemdash & 91.0 \\
		\bottomrule
	\end{tabular*}
\end{table}

With subgoal conditioning retained, removing KI reduces SR to 94.2\%, while removing LA distillation reduces it to 93.4\%; removing both yields 93.2\%. These results support aligning the policy-predicted latent action with the Stage-1 posterior and insulating the predictive path from action-loss gradients.

To distinguish inference-time subgoal conditioning from auxiliary supervision, we retain both future-prediction and LA distillation losses but remove the predicted subgoal from the action expert. This variant reaches 94.6\%, 1.8 percentage points below the reference without Stage-2 pretraining, supporting the benefit of explicitly using the decoded future for control. Removing LA distillation as well yields the \emph{Future Prediction} variant, which uses future prediction as auxiliary supervision, similar to VLA-JEPA and Mantis, and achieves 93.4\% SR. The \emph{LA Distillation Only} variant retains distillation without spatial future decoding and reaches 95.2\%, indicating additional value from the decoded spatial interface. Finally, Base VLA without LaWM, future-prediction supervision, or LA distillation reaches 91.0\%. Together, these comparisons support using the learned forward decoder as an inference-time interface in addition to predictive representation supervision.

Additional controls for model capacity and direct future-feature prediction are reported in Appendix~\ref{app:mechanism_ablations}.

\section{Limitations}
LaWAM is currently most effective in manipulation settings with relatively stable camera views. When camera motion dominates the observed transition, as in egocentric videos with abrupt shake or large viewpoint changes, LaWM can fail to learn a coherent latent action space. This limits the present formulation for humanoid or mobile robots whose observations are strongly shaped by self-motion. A second limitation is data coverage: fine-grained deformable-object dynamics, such as subtle cloth deformation during towel folding, are rare in the current training mixture, making them harder for LaWM to model reliably. Future work will scale LaWAM with broader data and model capacity to improve robustness under moving cameras and finer-grained physical interactions.

\section{Conclusion}
We presented LaWAM, a latent World-Action Model that brings predictive dynamics into robot policy inference without reconstructing pixel-space futures. LaWAM repurposes the forward decoder of a latent action model as LaWM, which expands policy-predicted latent actions into embodiment-grounded latent visual subgoals for action-chunk generation. Across simulated and real-world manipulation tasks, these subgoals provide action-relevant future context while avoiding the latency and parameter cost of pixel-space WAMs, suggesting that future prediction can serve as a compact latent interface between semantic instruction following and physically grounded control.

%%%%%%%%%%%%%%%%%%%%%%%%%%%%%%%%%%%%%%%%%%%%%%%%%%%%%%%%%%%%
%===============================================================================

\clearpage
\acknowledgments{This work was supported by the National Natural Science Foundation of China (Grant Nos. 92567204, 62472196, and 62406159), Zhongguancun Academy (Grant No. C20250301), and the Shenzhen Science and Technology Program (Grant No. AI2026016).}
\bibliography{references}  % .bib
\clearpage
\appendix

\section{Detailed RoboTwin Results}
\label{app:robotwin_detail}
\begin{table*}[t]
    \centering
    \tiny
    \caption{\textbf{Per-task success rates on RoboTwin under clean and randomized evaluation settings.}}
    \label{tab:robotwin_detail}
    \setlength{\tabcolsep}{2pt}
    \resizebox{\textwidth}{!}{
        \begin{tabular}{lcc|cc|cc|cc|cc|cc}
            \toprule
            Task                      & \multicolumn{2}{c}{Fast-WAM} & \multicolumn{2}{c}{GigaWorld-Policy} & \multicolumn{2}{c}{LingBot-VA} & \multicolumn{2}{c}{$\pi_{0.5}$} & \multicolumn{2}{c}{Motus} & \multicolumn{2}{c}{LaWAM}                                                                                             \\
            \cmidrule(lr){2-3}\cmidrule(lr){4-5}\cmidrule(lr){6-7}\cmidrule(lr){8-9}\cmidrule(lr){10-11}\cmidrule(lr){12-13}
                                      & Clean                        & Rand.                                & Clean                          & Rand.                           & Clean                     & Rand.                     & Clean        & Rand.        & Clean        & Rand.        & Clean          & Rand.        \\
            \midrule
            Adjust Bottle                   & \textbf{100}   & \textbf{100}   & \textbf{100}   & \textbf{100}   & 98             & 99             & \textbf{100}   & 99             & 89             & 93             & \textbf{100}   & \textbf{100}   \\
            Beat Block Hammer               & \textbf{99}    & \textbf{100}   & 86             & 86             & 96             & 98             & 96             & 93             & 95             & 88             & 90             & 93             \\
            Blocks Ranking RGB              & \textbf{100}   & \textbf{100}   & 92             & 96             & 98             & 96             & 92             & 85             & 99             & 97             & 97             & \textbf{100}   \\
            Blocks Ranking Size             & 94             & \textbf{98}    & 44             & 48             & \textbf{96}    & 89             & 49             & 26             & 75             & 63             & 93             & 89             \\
            Click Alarmclock                & \textbf{100}   & \textbf{100}   & \textbf{100}   & \textbf{100}   & \textbf{100}   & 99             & 98             & 89             & \textbf{100}   & \textbf{100}   & \textbf{100}   & \textbf{100}   \\
            Click Bell                      & \textbf{100}   & \textbf{100}   & \textbf{100}   & \textbf{100}   & \textbf{100}   & \textbf{100}   & 99             & 66             & \textbf{100}   & \textbf{100}   & \textbf{100}   & \textbf{100}   \\
            Dump Bin Bigbin                 & 95             & 93             & 92             & \textbf{100}   & 87             & 95             & 92             & 97             & 95             & 91             & \textbf{97}    & 95             \\
            Grab Roller                     & \textbf{100}   & \textbf{100}   & \textbf{100}   & \textbf{100}   & \textbf{100}   & \textbf{100}   & \textbf{100}   & \textbf{100}   & \textbf{100}   & \textbf{100}   & \textbf{100}   & \textbf{100}   \\
            Handover Block                  & 99             & 80             & 80             & 80             & \textbf{100}   & \textbf{91}    & 66             & 57             & 86             & 73             & 96             & 87             \\
            Handover Mic                    & \textbf{100}   & \textbf{100}   & 72             & 72             & 96             & 95             & 98             & 97             & 78             & 63             & 93             & 98             \\
            Hanging Mug                     & \textbf{65}    & \textbf{56}    & 16             & 12             & 21             & 33             & 18             & 17             & 38             & 38             & 51             & 43             \\
            Lift Pot                        & \textbf{100}   & \textbf{100}   & 98             & 98             & \textbf{100}   & \textbf{100}   & 96             & 85             & 96             & 99             & \textbf{100}   & 99             \\
            Move Can Pot                    & 95             & 92             & 76             & 78             & 93             & \textbf{93}    & 51             & 55             & 34             & 74             & \textbf{98}    & \textbf{93}    \\
            Move Pillbottle Pad             & \textbf{100}   & \textbf{100}   & 90             & 90             & \textbf{100}   & 99             & 84             & 61             & 93             & 96             & 97             & 90             \\
            Move Playingcard Away           & \textbf{100}   & \textbf{100}   & 78             & 72             & \textbf{100}   & \textbf{100}   & 96             & 84             & \textbf{100}   & 96             & \textbf{100}   & \textbf{100}   \\
            Move Stapler Pad                & 84             & 63             & 92             & 82             & 59             & 71             & 56             & 42             & 83             & 85             & \textbf{94}    & \textbf{87}    \\
            Open Laptop                     & \textbf{100}   & \textbf{100}   & 96             & 98             & 96             & 90             & 90             & 96             & 95             & 91             & \textbf{100}   & \textbf{100}   \\
            Open Microwave                  & 46             & 34             & 74             & 66             & 56             & 80             & 34             & 77             & \textbf{95}    & \textbf{91}    & 41             & 43             \\
            Pick Diverse Bottles            & 81             & 88             & 82             & 70             & \textbf{93}    & 87             & 81             & 71             & 90             & \textbf{91}    & 91             & 88             \\
            Pick Dual Bottles               & \textbf{100}   & 94             & 86             & 86             & \textbf{100}   & \textbf{100}   & 93             & 63             & 96             & 90             & \textbf{100}   & 95             \\
            Place A2B Left                  & 94             & 89             & 94             & 88             & 96             & \textbf{94}    & 87             & 82             & 88             & 79             & \textbf{98}    & 91             \\
            Place A2B Right                 & \textbf{98}    & 90             & 90             & 92             & 91             & \textbf{94}    & 87             & 84             & 91             & 87             & 89             & \textbf{94}    \\
            Place Bread Basket              & 86             & \textbf{94}    & 82             & 82             & \textbf{94}    & \textbf{94}    & 77             & 64             & 91             & \textbf{94}    & 92             & 85             \\
            Place Bread Skillet             & 92             & \textbf{94}    & 94             & 90             & \textbf{95}    & 92             & 85             & 66             & 86             & 83             & 90             & 83             \\
            Place Burger Fries              & \textbf{98}    & 92             & \textbf{98}    & 96             & 95             & \textbf{98}    & 94             & 87             & \textbf{98}    & \textbf{98}    & 93             & 96             \\
            Place Can Basket                & 72             & 67             & 78             & 74             & 83             & \textbf{80}    & 62             & 62             & 81             & 76             & \textbf{92}    & 65             \\
            Place Cans Plasticbox           & 98             & \textbf{100}   & \textbf{100}   & \textbf{100}   & \textbf{100}   & 98             & 94             & 84             & 98             & 94             & \textbf{100}   & 95             \\
            Place Container Plate           & 98             & \textbf{100}   & 98             & 96             & 97             & 98             & 99             & 95             & 98             & 99             & \textbf{100}   & \textbf{100}   \\
            Place Dual Shoes                & 88             & 88             & 96             & 84             & 94             & 81             & 75             & 75             & 93             & 87             & \textbf{98}    & \textbf{94}    \\
            Place Empty Cup                 & \textbf{100}   & \textbf{100}   & 90             & 90             & \textbf{100}   & \textbf{100}   & \textbf{100}   & 99             & 99             & 98             & 99             & \textbf{100}   \\
            Place Fan                       & \textbf{100}   & \textbf{96}    & 92             & 94             & 95             & 88             & 87             & 85             & 91             & 87             & 92             & 93             \\
            Place Mouse Pad                 & 84             & 89             & 88             & 90             & \textbf{92}    & \textbf{93}    & 60             & 39             & 66             & 68             & 91             & 84             \\
            Place Object Basket             & 82             & 90             & 90             & \textbf{92}    & 89             & 85             & 80             & 76             & 81             & 87             & \textbf{92}    & 90             \\
            Place Object Scale              & 83             & \textbf{88}    & 88             & 80             & \textbf{98}    & 87             & 86             & 80             & 88             & 85             & 95             & \textbf{88}    \\
            Place Object Stand              & 88             & 91             & \textbf{100}   & \textbf{98}    & 99             & 91             & 91             & 85             & 98             & 97             & 92             & 93             \\
            Place Phone Stand               & \textbf{98}    & \textbf{98}    & 82             & 72             & \textbf{98}    & 97             & 81             & 81             & 87             & 86             & 93             & 94             \\
            Place Shoe                      & 97             & 98             & 98             & 96             & 99             & 97             & 92             & 93             & 99             & 97             & \textbf{100}   & \textbf{100}   \\
            Press Stapler                   & \textbf{98}    & \textbf{100}   & 96             & 96             & 82             & 88             & 87             & 83             & 93             & 98             & \textbf{98}    & 97             \\
            Put Bottles Dustbin             & 93             & 82             & 72             & 70             & 82             & 87             & 84             & 79             & 81             & 79             & \textbf{94}    & \textbf{92}    \\
            Put Object Cabinet              & \textbf{94}    & \textbf{82}    & 74             & 74             & 87             & 80             & 80             & 79             & 88             & 71             & 90             & \textbf{82}    \\
            Rotate QRcode                   & \textbf{95}    & 94             & 90             & 84             & 92             & \textbf{95}    & 89             & 87             & 89             & 73             & 94             & 89             \\
            Scan Object                     & \textbf{96}    & 86             & 60             & 64             & 90             & 89             & 72             & 65             & 67             & 66             & \textbf{96}    & \textbf{90}    \\
            Shake Bottle                    & \textbf{100}   & \textbf{100}   & \textbf{100}   & \textbf{100}   & \textbf{100}   & 98             & 99             & 97             & \textbf{100}   & 97             & \textbf{100}   & \textbf{100}   \\
            Shake Bottle Horizontally       & \textbf{100}   & \textbf{100}   & \textbf{100}   & 98             & 99             & 99             & 99             & 99             & \textbf{100}   & 98             & \textbf{100}   & \textbf{100}   \\
            Stack Blocks Three              & 98             & 96             & 70             & 78             & \textbf{100}   & \textbf{99}    & 91             & 76             & 91             & 95             & 90             & 75             \\
            Stack Blocks Two                & \textbf{100}   & \textbf{100}   & \textbf{100}   & 94             & \textbf{100}   & 99             & 97             & \textbf{100}   & \textbf{100}   & 98             & \textbf{100}   & 97             \\
            Stack Bowls Three               & 77             & 86             & 70             & 72             & 89             & 79             & 77             & 71             & 79             & \textbf{87}    & \textbf{90}    & 80             \\
            Stack Bowls Two                 & 90             & 96             & 96             & 92             & 98             & \textbf{99}    & 95             & 96             & 98             & 98             & \textbf{100}   & \textbf{99}    \\
            Stamp Seal                      & 78             & 86             & 96             & \textbf{98}    & \textbf{98}    & 95             & 79             & 55             & 93             & 92             & 89             & 88             \\
            Turn Switch                     & 66             & 56             & 82             & \textbf{84}    & 54             & 57             & 62             & 54             & \textbf{84}    & 78             & 47             & 56             \\
            \midrule
            \textbf{Average}                & 91.98          & 90.52          & 86.36          & 85.04          & 91.50          & \textbf{90.92} & 82.74          & 76.76          & 88.66          & 87.02          & \textbf{92.64} & 89.80          \\
            \bottomrule
        \end{tabular}
    }
\end{table*}

%%%%%%%%%%%%%%%%%%%%%%%%%%%%%%%%%%%%%%%%%%%%%%%%%%%%%%%%%%%%

\section{Additional Mechanism Ablations}
\label{app:mechanism_ablations}
We further examine whether the gains arise from extra model capacity and whether directly predicting future features can replace the latent-action interface. As in the main component ablations, all settings in Table~\ref{tab:mechanism_ablations} omit Stage-2 pretraining and are evaluated on LIBERO-10. LaWAM without Stage-2 pretraining serves as the reference at 96.4\% SR.

\begin{table}[htbp]
    \centering
    \small
    \caption{\textbf{Capacity and predictive-interface controls on LIBERO-10.} All settings omit Stage-2 pretraining. Changes in SR are measured in percentage points relative to LaWAM without Stage-2 pretraining.}
    \label{tab:mechanism_ablations}
    \begin{tabular*}{\linewidth}{@{\extracolsep{\fill}}lcc@{}}
        \toprule
        Setting & SR ($\%$) $\uparrow$ & $\Delta$ SR (pp) \\
        \midrule
        LaWAM w/o Stage-2 Pretraining & \textbf{96.4} & --- \\
        Direct Future Prediction & 94.4 & $-2.0$ \\
        Capacity-Matched Adapter & 91.4 & $-5.0$ \\
        \bottomrule
    \end{tabular*}
\end{table}

\paragraph{Direct future prediction.}
This variant directly predicts the future DINO feature and uses it to condition action generation, bypassing the latent-action bottleneck. It achieves 94.4\% SR, 2.0 percentage points below the reference. This comparison supports the benefit of predicting through the intermediate latent action and the retained pretrained LAM decoder over direct future-feature prediction.

\paragraph{Capacity-matched adapter.}
This control retains the latent-to-subgoal conditioning path but reinitializes the decoder and trains it using only the action loss, without future-prediction or LA distillation supervision. KI is disabled, allowing action-loss gradients to pass through the subgoal-conditioning path and update the reinitialized decoder. It achieves 91.4\% SR, compared with 91.0\% for Base VLA. The pretrained predictive interface improves SR by 5.0 percentage points over this capacity-matched control.

\section{Additional LaWM Rollout Analysis}
\label{app:lawm_analysis}

\begin{figure}[t]
	\centering
	\includegraphics[width=0.92\linewidth]{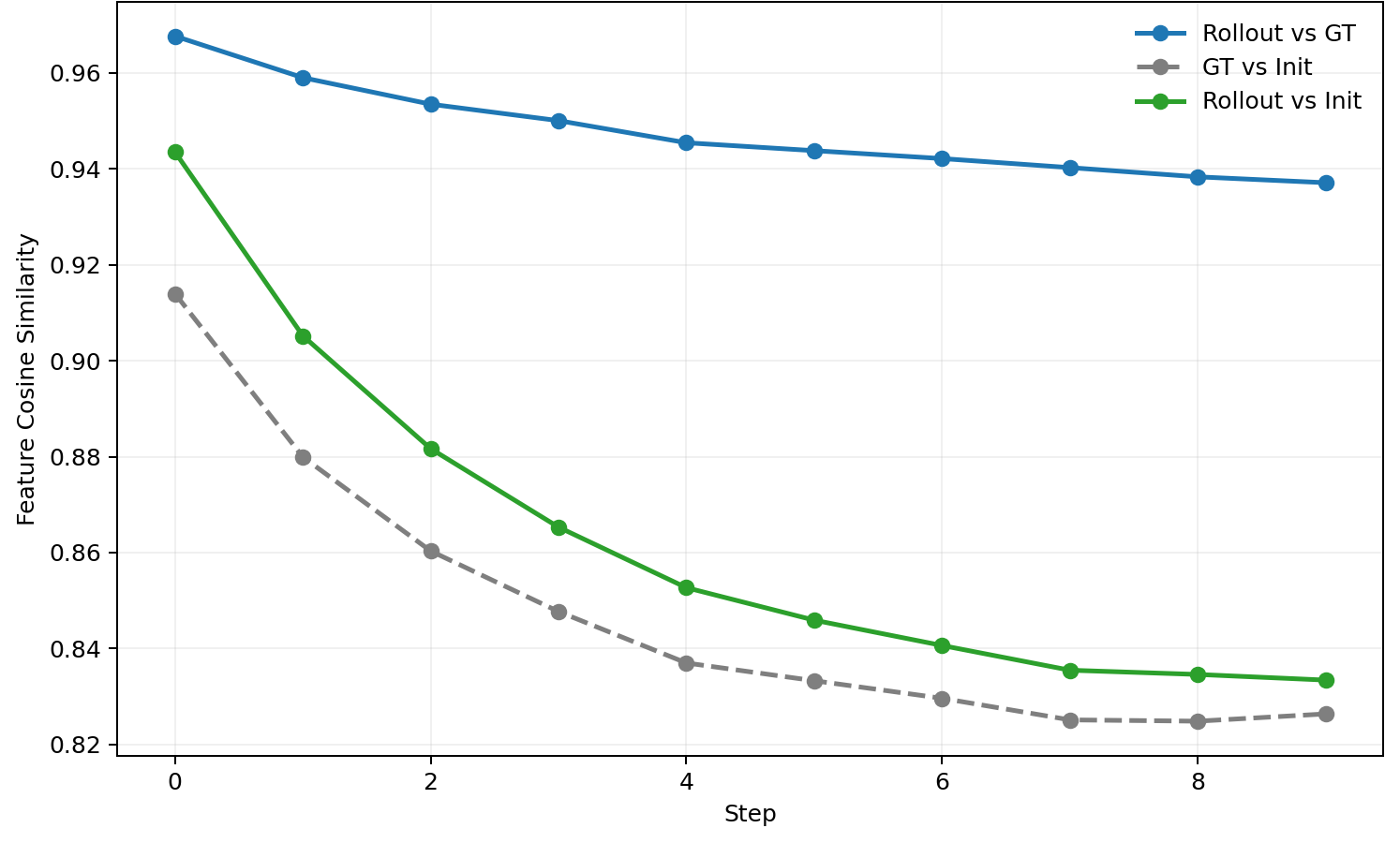}
	\caption{\textbf{Average LaWM rollout results over 500 LIBERO trajectories.} Curves show feature cosine similarity for rollout vs. ground-truth future states (blue), ground-truth future vs. initial states (gray), and rollout vs. initial states (green). The consistently high rollout--ground-truth similarity, together with the decreasing similarity to the initial state, indicates that LaWM follows the true latent dynamics rather than remaining close to the starting observation.}
	\label{fig:lawm_rollout_error}
\end{figure}

We use several complementary visualizations to examine the dynamics modeled by LaWM. Fig.~\ref{fig:lawm_rollout_error} aggregates open-loop rollout behavior over 500 LIBERO trajectories by comparing the predicted horizon features against ground-truth future features and the initial features. The rollout remains close to the true future state while moving away from the initial state, indicating that LaWM models nontrivial latent dynamics rather than simply preserving the current observation. This aggregate probe supports LaWM's dynamics-modeling ability over extended open-loop rollouts.

\begin{figure}[t]
	\centering
	\includegraphics[width=\linewidth]{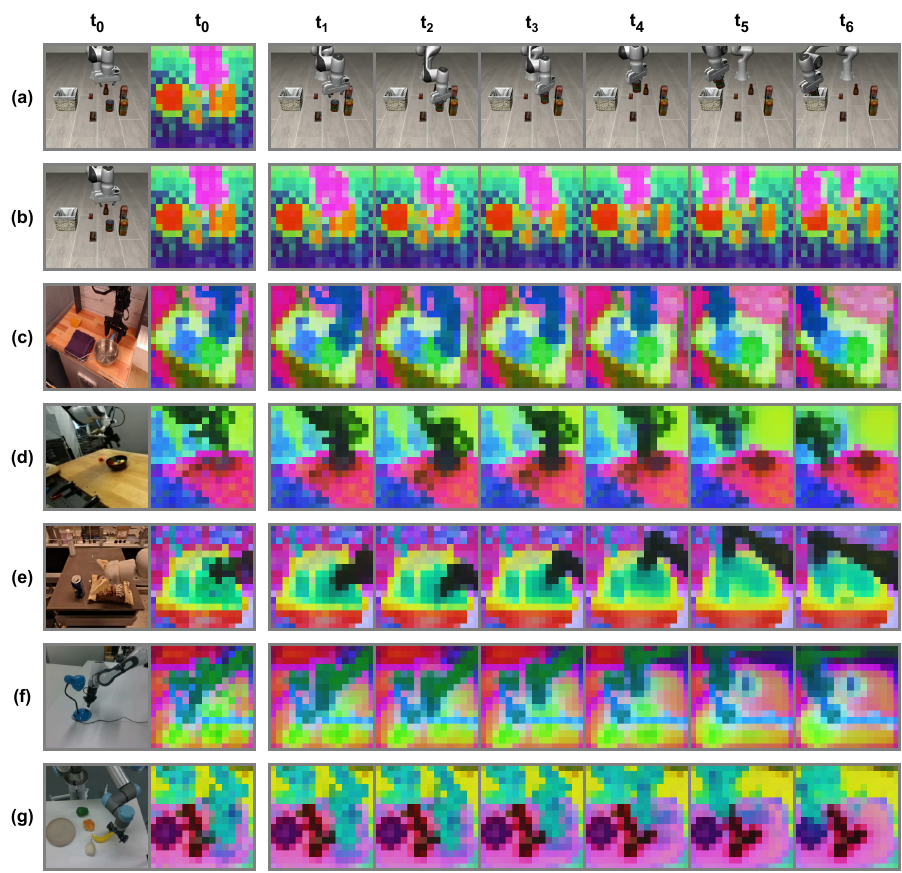}
	\caption{\textbf{Single-arm cross-embodiment rollout.}}
	\label{fig:lam_single_pca}
\end{figure}

\begin{figure}[t]
	\centering
	\includegraphics[width=\linewidth]{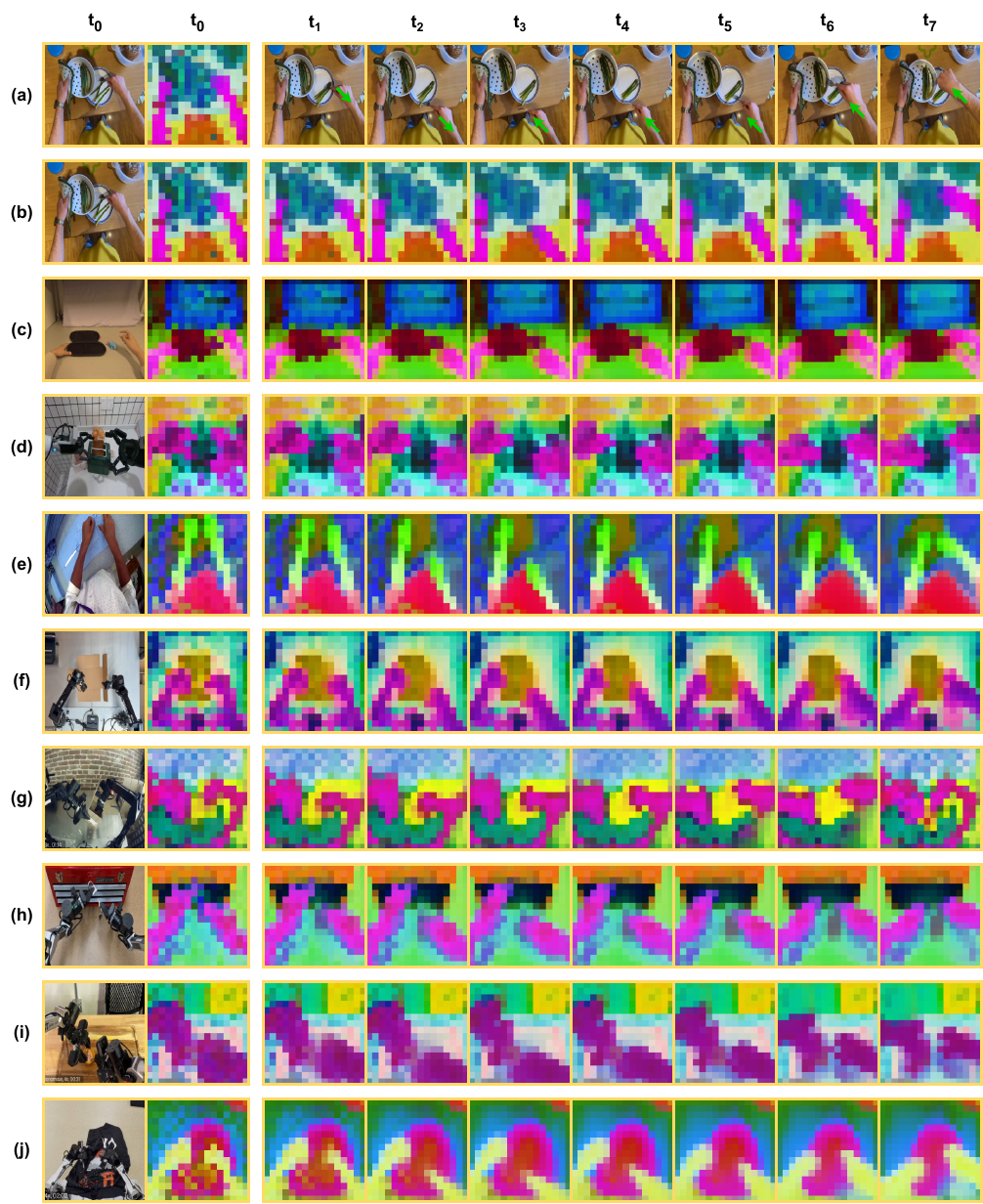}
	\caption{\textbf{Dual-arm cross-embodiment rollout.} Panels (e)--(j) show unseen scenes; panels (f)--(j) use screenshots from \emph{pi.website}.}
	\label{fig:lam_dual_pca}
\end{figure}

Figs.~\ref{fig:lam_single_pca} and~\ref{fig:lam_dual_pca} provide additional cross-environment and cross-embodiment rollout examples. These visualizations complement Fig.~\ref{fig:lam_intro_pca}: applying the same latent action in different visual contexts yields coherent but context-specific latent changes. We interpret this as evidence that LaWM uses the current latent visual state to ground latent actions in the embodiment and scene, while the latent action specifies the abstract transition to be realized.

\section{Qualitative Visualization Protocol}
\label{app:visualization}
\begin{figure}[t]
	\centering
	\includegraphics[width=\linewidth]{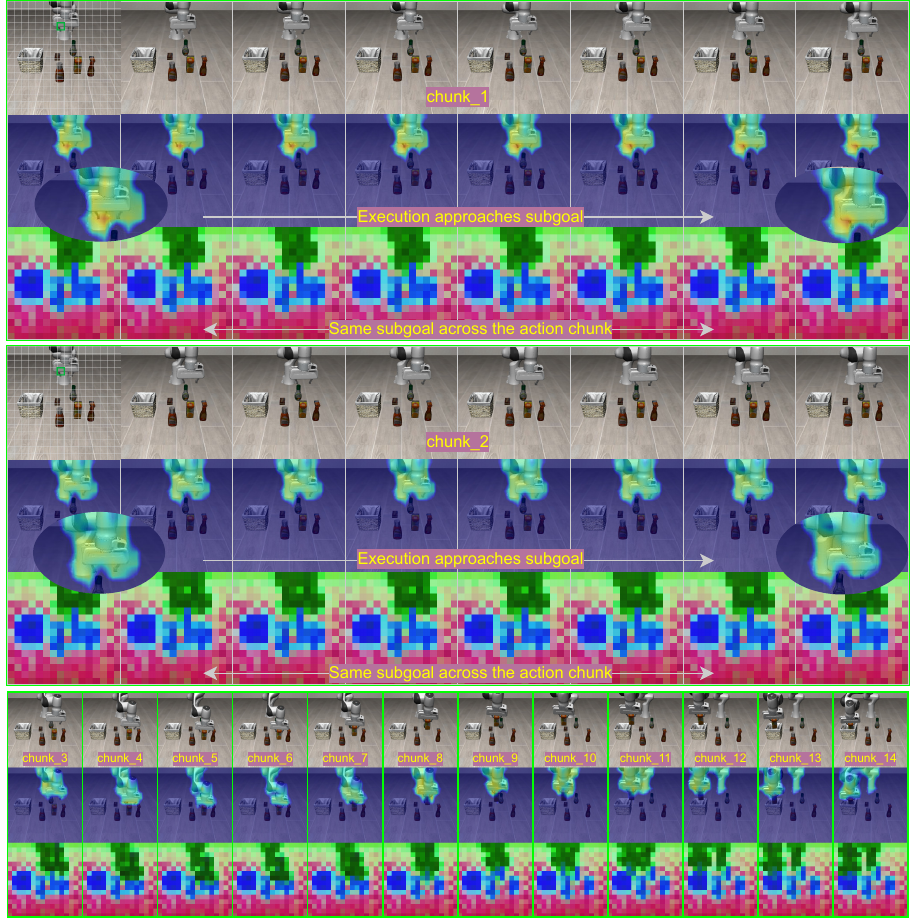}
	\caption{\textbf{Qualitative LIBERO examples showing LaWAM action execution together with latent world rollouts.} The predicted latent subgoals provide compact future dynamics that guide action generation without iterative pixel-space video prediction.}
	\label{fig:lawam_libero_full}
\end{figure}

Figs.~\ref{fig:libero_chunk}, \ref{fig:lawam_libero_full}, and~\ref{fig:pp_duck}--\ref{fig:clean_table_subgoals} visualize predicted LaWM subgoals in the latent DINO feature space. The real-world examples in Figs.~\ref{fig:pp_duck}--\ref{fig:clean_table_subgoals} share a common format: RGB observations at chunk endpoints, future-feature similarity overlays, and PCA projections of predicted subgoals.

For each sequence, we select a robot-arm patch in the initial observation, extract its DINO feature, and measure its cosine similarity to every patch in the predicted subgoal feature map. The green box in the first panel of Fig.~\ref{fig:lawam_libero_full} shows one selected patch. The resulting heatmap indicates the predicted location of the selected arm feature, while PCA summarizes the spatial structure of the predicted features. For the robot executions in Figs.~\ref{fig:pp_duck}--\ref{fig:fold_towel_all}, comparing the heatmap with the observed arm at the end of each chunk illustrates alignment between the predicted subgoal and executed motion.

Figure~\ref{fig:clean_table_subgoals} applies the same visualization to offline replay of a Clean Table training demonstration. LaWAM receives recorded observations every 24 frames (1.2\,s), and the RGB panels show the final frame before the next input. The similarity overlays use a fixed arm token from the replay segment's initial observation, and the PCA projections share a common basis. The 28 displayed chunks are sampled every 6--7 chunks across the sequence.

% \clearpage
\section{Architecture and Implementation Details}
\label{app:arch_details}

\subsection{Comparison of Predictive Interfaces}
\label{app:predictive_interfaces}
\begin{figure}[H]
    \centering
    \includegraphics[width=\linewidth]{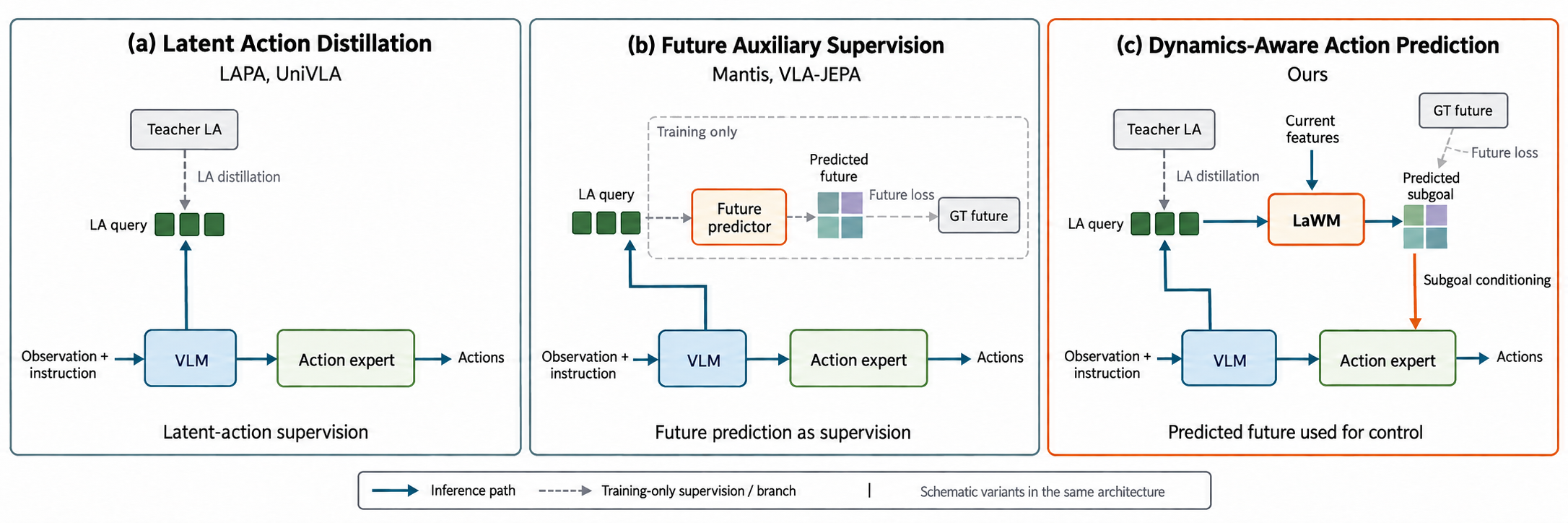}
    \caption{\textbf{From auxiliary supervision to future-conditioned action generation.} (a) Latent-action supervision, exemplified by LAPA and UniVLA~\cite{ye2025latentactionpretrainingvideos,bu2025univlalearningacttaskcentric}. (b) Future-prediction supervision, exemplified by Mantis and VLA-JEPA~\cite{mantisVersatileVisionlanguageaction2025,vlajepaEnhancingVisionlanguageaction2026}: the VLM-to-action-expert path includes action-query representations that attend to the learned queries within the VLM; only the future predictor is omitted at inference. (c) LaWAM retains LaWM at inference and explicitly conditions action generation on its predicted spatial subgoal. Solid arrows denote inference paths; dashed arrows denote training-only supervision or branches.}
    \label{fig:predictive_interfaces}
\end{figure}

As illustrated in Fig.~\ref{fig:predictive_interfaces}, auxiliary future supervision and explicit future conditioning differ in how the decoded prediction enters action generation. In Mantis and VLA-JEPA~\cite{mantisVersatileVisionlanguageaction2025,vlajepaEnhancingVisionlanguageaction2026}, learned queries condition a future predictor trained against future visual targets. Action queries attend to these learned query representations within the VLM to form the conditioning input to the action expert, rather than using the decoded future predictions. Future supervision thus enriches the query representations, while LaWAM additionally passes the decoded spatial future to the action expert at inference. This directly reuses the learned forward mapping to provide spatial transition information for control.

\subsection{Latent World Model Architecture}
\label{app:lawm_arch}
LaWM operates on features from the distilled DINOv3 ViT-B/16 encoder \cite{simeoni2025dinov3}. Its inverse-dynamics encoder and decoder are both 24-layer transformer modules. The encoder follows a V-JEPA2-style spatiotemporal design \cite{assran2025vjepa2}: visual patches from the current and horizon observations are flattened into a single token sequence and processed jointly to infer the latent action posterior. Unlike the additive token injection used in Genie~\cite{bruce2024geniegenerativeinteractiveenvironments}, our decoder conditions on the latent action $z$ through adaptive layer normalization, which we found more stable in the cross-embodiment setting; additive injection can make fluctuations in the latent-action norm induce global shifts of visual tokens and cause sharp loss spikes. The auxiliary state predictor uses lightweight MLP heads to handle inconsistent state semantics across robot embodiments and is discarded after LaWM training.

\subsection{Latent World Action Model Architecture}
\label{app:lawam_arch}
This section provides the architecture specifications for the policy described in Sec.~\ref{sec:lawam}. LaWAM follows the Qwen-GR00T architecture. We use the first 16 transformer layers of Qwen3-VL as the VLM backbone, and instantiate the action expert with four Alternate-DiT blocks, corresponding to 16 transformer layers in total. The hidden dimension is 1024.

The policy input sequence contains the primary-view observation, the task instruction, latent-action query tokens, optional auxiliary-view images, and action-query tokens. All non-query tokens serve as context. With this ordering and a causal attention mask, the latent-action queries gather the information needed to drive LaWM, while the action queries still attend to the full semantic context used for control. The semantic conditioning stream uses the full VLM hidden state; the dynamics stream uses the current and predicted spatial features $(u,\hat{u}_T)$.

All robot action labels are converted to an end-effector (EEF) representation. During policy training and evaluation, we do not provide proprioceptive state inputs, which helps avoid overfitting to trajectory-specific state traces and improves spatial generalization \cite{zhao2025needproprioceptivestatesvisuomotor}. LaWAM uses RGB-only observations, with images resized to $256\times256$.

\subsection{Physical-Time Alignment for Mixed-Frequency Data}
\label{app:physical_time}
Stage-2 training combines robot datasets with different native control frequencies. We retain these frequencies while aligning action chunks and visual subgoals over a fixed physical interval $\tau$. For a sample from a dataset with control frequency $h$, the action horizon and token timestamps are
\begin{equation}
    \label{eq:physical_time_alignment}
    T=\mathrm{round}(\tau h),\qquad t_i=i/h,\quad i=1,\ldots,T.
\end{equation}
Here $T$ is the action-chunk length and varies with the sample's native control frequency. Each sample contains the valid action sequence $a_{1:T}$ paired with the horizon observation at elapsed time $\tau$. For example, under the same $\tau$, a 5\,Hz dataset uses fewer action tokens than a 20\,Hz dataset, while both predict the visual state reached after the same physical duration. Encoding tokens by $t_i$, rather than their discrete indices alone, aligns actions with the predicted subgoal across datasets.

We implement the physical-time encoding of timestamp $t_i$ using sinusoidal features:
\begin{equation}
	\operatorname{PE}(t_i)=\mathrm{Concat}\left[\sin(t_i\nu_k),\cos(t_i\nu_k)\right]_{k=0}^{K-1},
\end{equation}
where $\nu_k=\exp\!\left(-\log(P_{\max})\cdot \frac{k}{\max(K-1,1)}\right)$ and $K=\lceil d/2\rceil$. Here $d$ is the action-query embedding dimension, and $P_{\max}$ sets the range of sinusoidal frequencies. We add $\operatorname{PE}(t_i)$ to the corresponding action-query embedding before action denoising. During batching, variable-length chunks are padded and masked, while valid tokens retain their native timestamps. Tokens with the same elapsed time receive the same temporal code even if they occupy different indices. The controlled mixed-frequency benchmark construction is described in Sec.~\ref{app:mixed_frequency}.

\subsection{Mixed-Frequency Training Experiment}
\label{app:mixed_frequency}
We evaluate physical-time encoding in a controlled LIBERO experiment. Starting from the same native 20\,Hz demonstrations, we create 10\,Hz and 5\,Hz versions by temporal downsampling and train LaWAM from scratch on the combined 5/10/20\,Hz data. The branches share the source demonstrations, task distribution, embodiment, visual domain, and language instructions; no additional expert trajectories are introduced. A model trained from scratch on the 20\,Hz data alone serves as the native-rate reference.

\begin{figure}[t]
	\centering
	\includegraphics[width=0.62\linewidth]{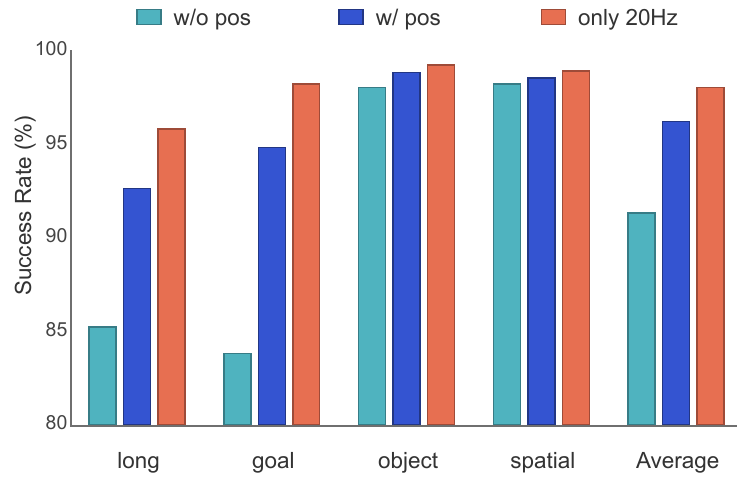}
	\caption{\textbf{Mixed-frequency training results on LIBERO.} Joint training across 5/10/20\,Hz versions downsampled from the same LIBERO data shows that physical-time encoding resolves control-frequency confusion.}
	\label{fig:mix_hz}
\end{figure}

As shown in Fig.~\ref{fig:mix_hz}, mixed-frequency training without physical-time encoding substantially reduces success relative to the native-rate reference, while adding the encoding largely recovers performance. This supports using elapsed physical time to align action tokens across control frequencies.

\subsection{Training Details}
\label{app:training_details}
\paragraph{LaWM training.}
For LaWM training, we use continuous latent actions and optimize Eq.~\ref{eq:lam_objective} with 16 H100 GPUs for 100k steps using AdamW, learning rate $3\times10^{-4}$, weight decay $10^{-2}$, and a global batch size of 1024. We set the KL regularization weight to $\beta=10^{-5}$. The fixed physical-time horizon is 1.2\,s for robot teleoperation videos and 0.4\,s for egocentric human videos. Before DINOv3 encoding, we apply different random crops and color augmentations to encoder and decoder views, while keeping augmentations temporally consistent within each encoder clip; this discourages LaWM from memorizing embodiment-specific pixel layouts.

\paragraph{Stage-2 initialization.}
We initialize the VLM backbone from pretrained Qwen3-VL weights and retain both components of the Stage-1 latent action model. The posterior encoder is frozen and supplies teacher latent actions, while its forward decoder serves as a trainable LaWM. The DINOv3 visual encoder remains frozen. The latent-action projector and Alternate-DiT action expert are initialized from scratch.

\paragraph{Supervision and gradient routing.}
For each training sample, the frozen visual encoder extracts $(u,u_T)$ and the frozen posterior encoder provides the teacher latent action $z$. In Eq.~\ref{eq:lawam_objective}, expectations are taken over training samples and, for distillation, the teacher latent actions. The superscript $(2)$ distinguishes future prediction driven by the policy prior from the posterior-driven prediction in Stage 1. KI applies stop-gradient to $\hat{u}_T$ before action conditioning, preventing the action loss from updating LaWM through this branch. LaWM remains trainable through future-feature supervision, and distillation continues to supervise the policy prior. The action loss continues to train the action expert and its semantic conditioning pathway.

\paragraph{LaWAM policy training and evaluation.}
Stage 2 uses robot trajectories with language instructions; egocentric human videos contribute through the Stage-1 dynamics prior but are not used for policy integration. For the joint objective in Eq.~\ref{eq:lawam_objective}, we set $\lambda_{\mathrm{distill}}=\lambda_{\mathrm{wm}}=0.1$ in all experiments. We run Stage-2 policy pretraining for 200k steps on 64 H100 GPUs with a global batch size of 1024. The action expert uses a learning rate of $10^{-4}$, while the other trainable modules use $3\times10^{-5}$.

For post-training on each benchmark, we use a learning rate of $10^{-4}$ with cosine decay. Variants without Stage-2 pretraining skip the 200k-step policy-pretraining run and proceed to benchmark-specific post-training. Benchmark-specific training schedules and evaluation protocols are provided in Appendix~\ref{app:exp_details}.

At deployment, the posterior encoder is discarded. The policy predicts $\hat{z}$ from the current observation and instruction, LaWM produces one latent subgoal in a single forward pass, and the action expert generates the corresponding action chunk. Unless otherwise specified, all policies are evaluated with 10 denoising steps. We compare against recent state-of-the-art VLA, latent-action, and WAM baselines, using benchmark success rates from the original papers whenever available and the strongest reproduced results otherwise. The inference-efficiency measurement protocol is detailed in Appendix~\ref{app:efficiency_protocol}.

\section{Experimental Details}
\label{app:exp_details}

\subsection{Inference Efficiency and Training Cost}
\label{app:efficiency_protocol}
\paragraph{Shared measurement conditions.}
For the LIBERO latency comparison, we measure all methods on the same A100 GPU using BF16 precision, 10 denoising steps, and matched batch size and kernel/compilation settings. We run 1,000 repeated action-chunk predictions with CUDA-synchronized timing and report the mean model-only wall-clock latency per action chunk, excluding simulator and robot execution overhead. LaWM predicts each latent subgoal in a single forward pass; the 10 denoising steps in LaWAM apply to the action expert.

\paragraph{Peak memory.}
Measured on the same A100 in BF16 without batching, peak memory is 6.04\,GB for LaWAM and 23.68\,GB for Fast-WAM.

\paragraph{Native benchmark configurations.}
Each method retains its official benchmark task configuration, including input resolution and prediction horizon, to reproduce the corresponding benchmark success rates. No additional inference-acceleration techniques are applied beyond those present in each method's official implementation. The comparison therefore uses shared measurement conditions while preserving each method's native resolution, horizon, and computational workload.

\paragraph{Parameter accounting.}
For WAM parameter counts reported in the paper, we exclude the video-diffusion VAE and the text encoder, which can contain more than 5B parameters. This parameter-count convention is separate from the model-only latency measurement described above.

\paragraph{Training cost.}
Table~\ref{tab:training_cost} reports LaWAM training costs in H100 GPU-hours, separating Stage-1 LaWM pretraining, Stage-2 policy pretraining, and benchmark-specific post-training.

\begin{table}[htbp]
    \centering
    \small
    \caption{\textbf{LaWAM training cost by stage.} LIBERO and RoboTwin post-training costs are reported separately.}
    \label{tab:training_cost}
    \begin{tabular*}{\linewidth}{@{\extracolsep{\fill}}lr@{}}
        \toprule
        Training stage & H100 GPU-hours \\
        \midrule
        Stage-1 LaWM pretraining & 400 \\
        Stage-2 policy pretraining & 3,000 \\
        LIBERO post-training & 40 \\
        RoboTwin post-training & 1,200 \\
        \bottomrule
    \end{tabular*}
\end{table}

\subsection{LIBERO Protocol}
\label{app:libero_protocol}
We follow the standard LIBERO protocol \cite{liu2023libero} and train on four suites: LIBERO-Spatial, LIBERO-Object, LIBERO-Goal, and LIBERO-Long. Following the OpenVLA evaluation setup \cite{kimOpenVLAOFTFinetuningVisionlanguageaction2025}, we remove failed demonstrations from the training data. LaWAM is trained for 25k steps with a global batch size of 256. We report success rates over 2,000 trials across 40 tasks.

\subsection{RoboTwin Protocol}
\label{app:robotwin_protocol}
We follow the multi-task training setup used by Motus \cite{bi2025motusunifiedlatentaction} and LingBot-VA \cite{li2026causalworldmodelingrobot}: models are trained on a mixture of 2,500 demonstrations collected in clean scenes and 25,000 demonstrations collected under heavy scene randomization, spanning over 50 tasks. LaWAM is trained for 100k steps with a global batch size of 1024, requiring approximately 1,200 H100 GPU-hours. We report average success rates over 100 trials per task under both clean and randomized settings.

\begin{figure}[t]
	\centering
	\begin{subfigure}[t]{0.48\linewidth}
		\centering
		\includegraphics[width=\linewidth]{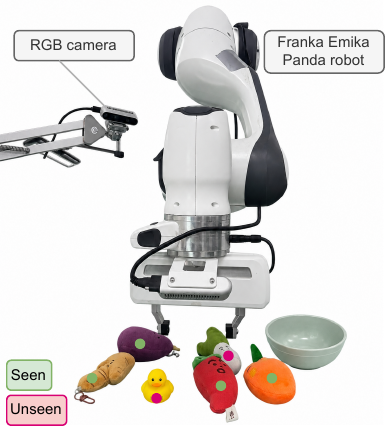}
		\caption{Franka setup.}
	\end{subfigure}
	\hfill
	\begin{subfigure}[t]{0.48\linewidth}
		\centering
		\includegraphics[width=\linewidth]{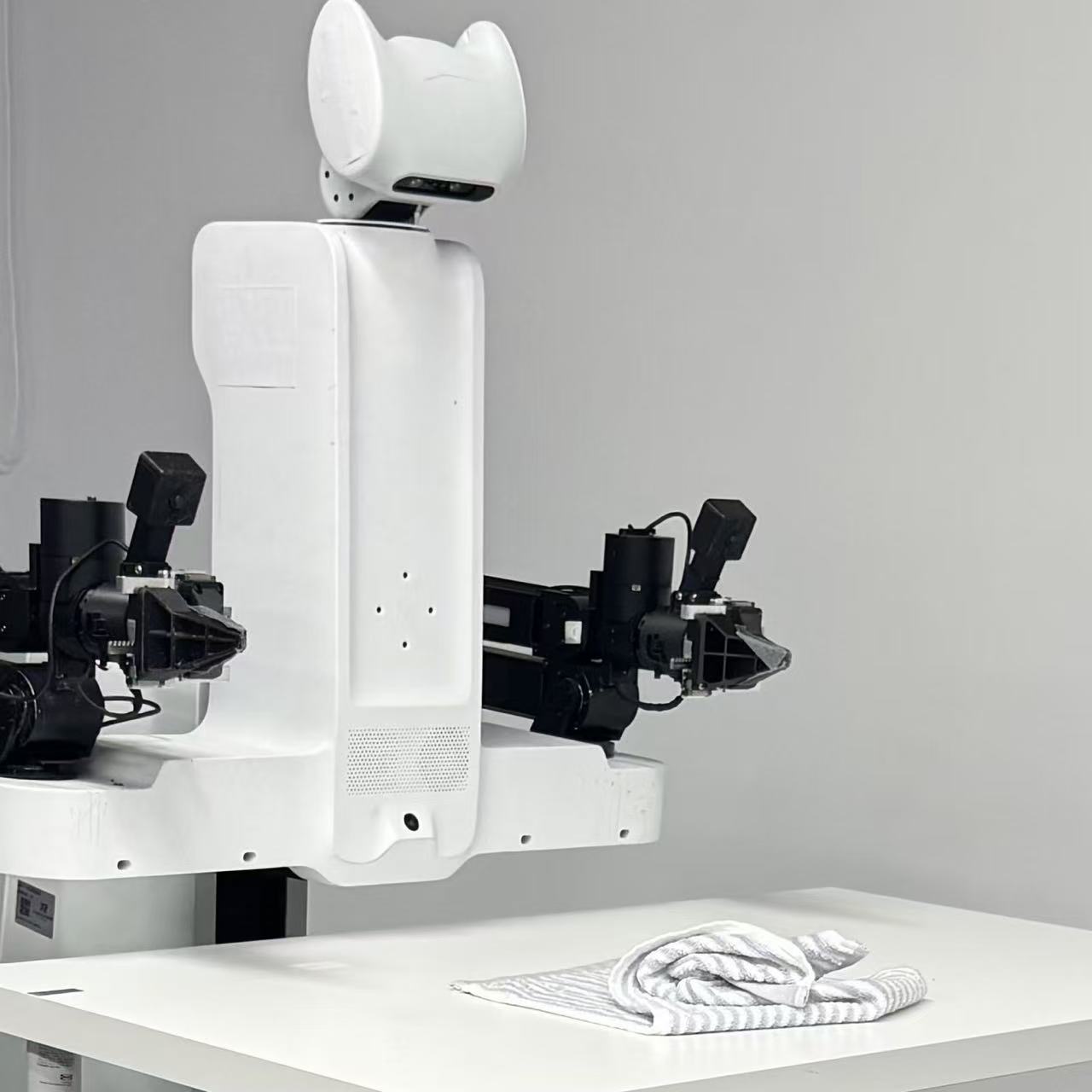}
		\caption{Quanta X1 setup.}
	\end{subfigure}
	\caption{\textbf{Real-world robot setups.} (a) A Franka Emika Panda robot with an external RGB camera is used for the single-arm pick-and-place and drawer-opening tasks. (b) The Quanta X1 bimanual robot is used for towel folding and table clearing.}
	\label{fig:real_world_setups}
\end{figure}

\subsection{Real-World Protocol}
\label{app:real_world_protocol}
We use two physical robot platforms, shown in Appendix Fig.~\ref{fig:real_world_setups}. Pick-and-place and drawer opening are conducted on a Franka Emika Panda arm with a parallel gripper and an external RGB camera observing the workspace, while towel folding and table clearing use a Quanta X1 bimanual robot. We use 150 real-world demonstrations for each Franka task, 280 for towel folding, and 302 for table clearing. All compared methods are fine-tuned on the same task-specific demonstrations with matched fine-tuning hyperparameters.

\paragraph{Evaluation protocol.}
Pick-and-Place, Drawer Opening, and Towel Folding in Table~\ref{tab:real_world_success} each use 30 trials spanning seen and unseen initial conditions. Their initial configurations are generated once and fixed across methods. Clean Table, Clean Table--Light, and Fold Towel--Light in the same table each use 50 trials per method and condition. We report the fraction of successful trials, with 95\% Wilson binomial confidence intervals for the 50-trial evaluations in Table~\ref{tab:real_world_confidence}.

\begin{table}[htbp]
    \centering
    \small
    \caption{\textbf{Confidence intervals for the 50-trial real-world evaluations.} SR (\%) with 95\% Wilson confidence intervals in brackets. Each method is evaluated over 50 trials per condition.}
    \label{tab:real_world_confidence}
    \begin{tabular*}{\linewidth}{@{\extracolsep{\fill}}lccc@{}}
        \toprule
        Method & Fold Towel--Light & Clean Table & Clean Table--Light \\
        \midrule
        $\pi_{0.5}$ & 46 [33.0, 59.6] & 28 [17.5, 41.7] & 10 [4.3, 21.4] \\
        Fast-WAM & 14 [7.0, 26.2] & 6 [2.1, 16.2] & 0 [0.0, 7.1] \\
        LaWAM & \textbf{68} [54.2, 79.2] & \textbf{42} [29.4, 55.8] & \textbf{24} [14.3, 37.4] \\
        \bottomrule
    \end{tabular*}
\end{table}

\paragraph{Pick-and-place.}
The robot must grasp a target object and place it into a bowl. The test objects include plush vegetable toys and toy ducks, covering irregular shapes and different surface properties; in particular, the duck is made of smooth plastic, which creates a more challenging contact condition for stable grasping, and the toy scallion is longer than the bowl diameter, requiring accurate placement despite the size mismatch. Across the object set, we evaluate 30 randomly generated initial configurations that vary the object location, bowl location, and object pose. Some test positions slightly extend beyond the spatial region covered by the training demonstrations, providing a controlled probe of spatial generalization. A qualitative pick-and-place execution with LaWM subgoal visualizations is shown in Appendix Fig.~\ref{fig:pp_duck}.

\paragraph{Drawer opening.}
The robot must grasp the handle of a closed drawer and pull it open. Due to the drawer material and sliding contact, pulling in a direction that is not parallel to the drawer motion creates substantial resistance, making the task sensitive to the grasp pose and pull direction. The drawer front edge is randomized within a predefined placement region, and the drawer is initialized with its orientation parallel to the short edge of the table before applying a random rotational perturbation of up to $15^\circ$. We evaluate 30 trials that uniformly cover 10 predefined initial configurations for this task. A representative drawer-opening rollout is visualized in Appendix Fig.~\ref{fig:opendrawer}.

\paragraph{Towel folding.}
The Quanta X1 operates on a deformable towel and must complete a long-horizon folding sequence. The task begins by shaking out and flattening the towel, then requires two long-edge folds followed by one short-edge fold. Even an experienced teleoperator requires about 70\,s to complete the full sequence, highlighting the task's long-horizon nature. This setting stresses bimanual coordination, contact-rich cloth manipulation, and temporal consistency over a substantially longer horizon than the rigid-object and articulated-object tasks. The full towel-folding sequence is shown in Appendix Fig.~\ref{fig:fold_towel_all}.

\begin{figure}[p]
    \centering
    \includegraphics[width=\linewidth,height=0.84\textheight,keepaspectratio]{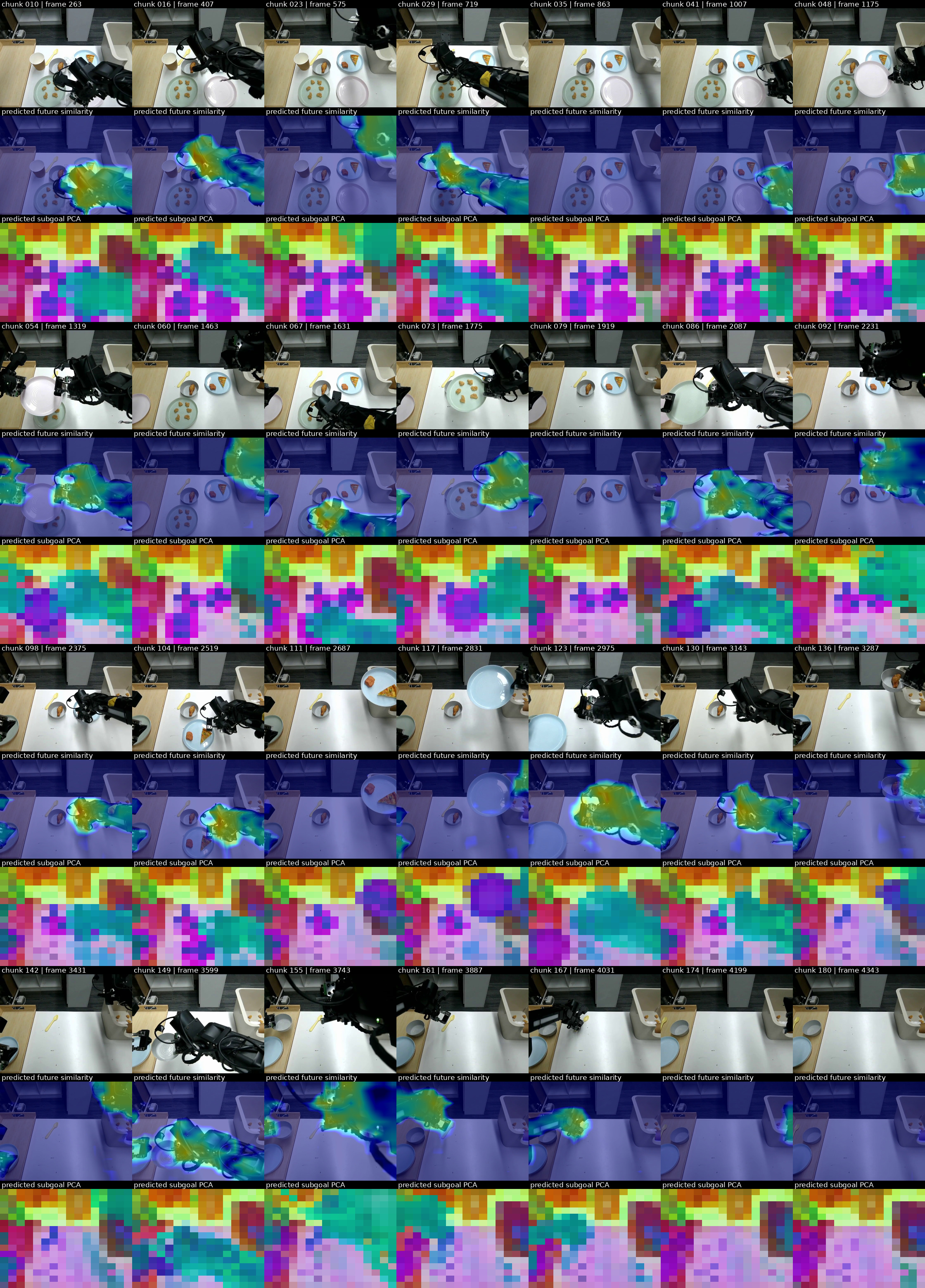}
    \caption{\textbf{LaWM subgoal visualization for Clean Table.} Offline replay of a training demonstration on Quanta X1, showing 28 sampled chunks across the table-clearing sequence.}
    \label{fig:clean_table_subgoals}
\end{figure}

\paragraph{Clean Table.}
The robot clears a table through eight object-level subtasks involving two tissues, a paper cup, three plates, a bowl, and a spoon (Fig.~\ref{fig:real_world_rollouts}(d)). The sequence combines grasping, emptying contents into a bin, bimanual handoffs, and placing objects on a collection tray. All 302 teleoperated demonstrations are used for fine-tuning. A trial is successful only if all eight subtasks are completed. Figure~\ref{fig:clean_table_subgoals} visualizes LaWAM's chunk-level future predictions during offline replay of a training demonstration.

\paragraph{Lighting perturbations.}
For each method, Clean Table and Clean Table--Light share one task-specific checkpoint; Fold Towel and Fold Towel--Light likewise share one checkpoint. These checkpoints are trained on normal-lighting demonstrations with color augmentation enabled. The Light suffix indicates an additional red, green, or purple lamp illuminating the workspace during evaluation, changing object and background appearance. Model weights and task success criteria are unchanged between the two lighting conditions; trial counts follow the evaluation protocol above.

\paragraph{Failure inspection.}
For Fast-WAM, failures on the two Franka tasks are often associated with spatial perception near the target, such as inaccurate object-distance estimates before grasping in pick-and-place and handle-distance estimates in drawer opening. For LingBot-VA, towel-folding failures are closely tied to high inference latency: while the policy is generating the next action, the towel can continue moving and remain wrinkled, making it difficult to shake out folds promptly and causing the generated action to be conditioned on a stale visual state. LaWAM still shows a limitation in deformable-object manipulation, where the current LaWM feature resolution is not yet fine-grained enough to precisely capture subtle cloth deformations. Nevertheless, by accurately predicting robot-arm subgoals, LaWAM achieves the strongest towel-folding performance among the compared methods.

% \begin{figure}[t]
% 	\centering
% 	\includegraphics[width=\linewidth]{picture/token_pca_timeline.pdf}
% 	\caption{Open-loop LaWM rollout from the initial observation over a full LIBERO trajectory. Starting from observed frames, LaWM rolls out future latent visual states, which are projected with PCA and overlaid on the target observations. From top to bottom: observations, predicted latent rollout visualized by PCA, PCA overlay on target observation, and end-effector state trajectories. Solid curves denote ground-truth states and dashed curves denote states predicted from the rollout latent actions, showing that LaWM preserves temporally consistent dynamics over long horizons.}
% 	\label{fig:pca_timeline}
% \end{figure}

\begin{figure}[t]
	\centering
	\includegraphics[width=\linewidth]{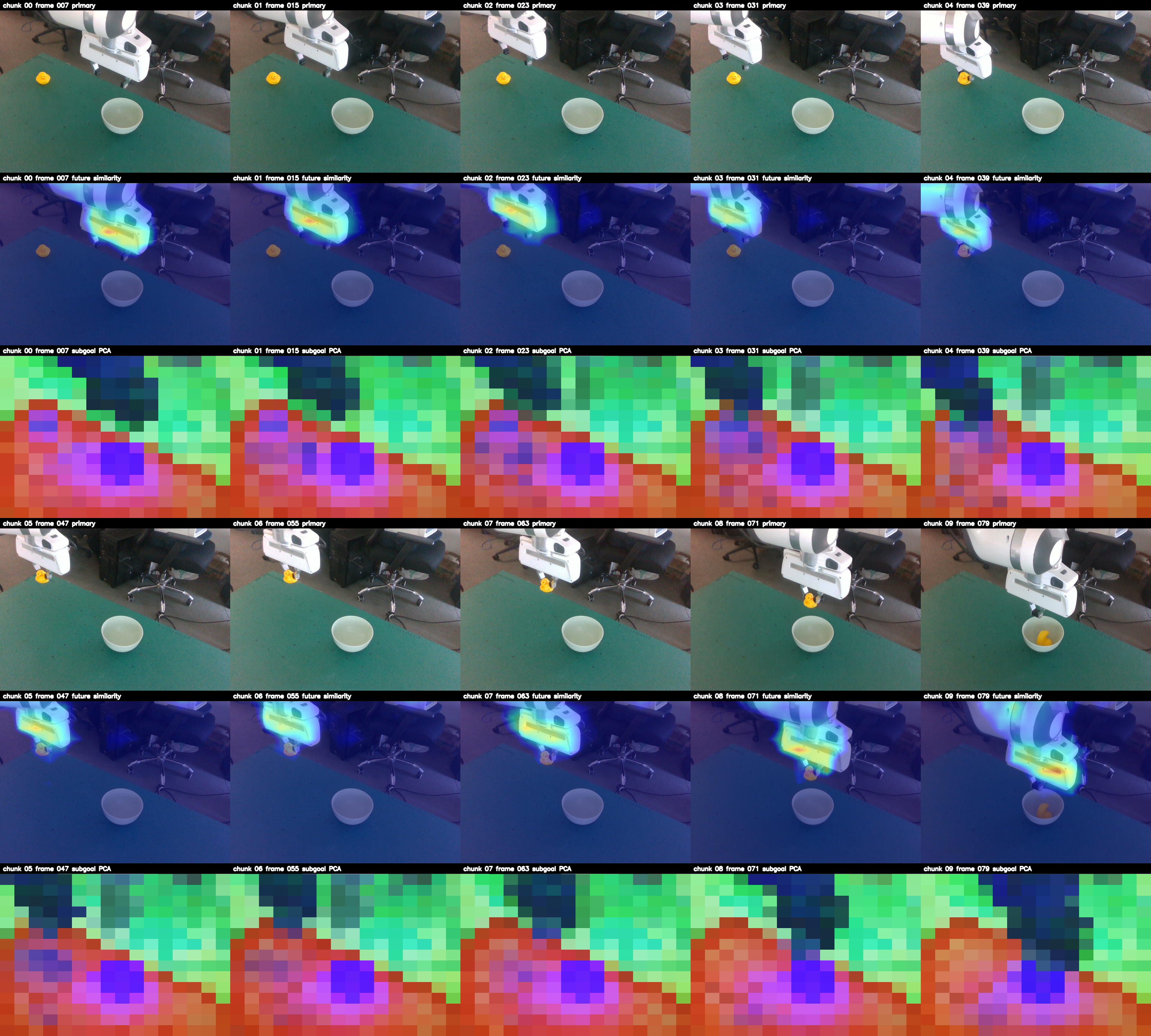}
	\caption{\textbf{LaWM subgoal visualization for Pick-and-Place.} The Franka arm approaches, grasps, and places a toy duck into the bowl. Visualization conventions for Figs.~\ref{fig:pp_duck}--\ref{fig:clean_table_subgoals} are described in Appendix~\ref{app:visualization}.}
	\label{fig:pp_duck}
\end{figure}

\begin{figure}[t]
	\centering
	\includegraphics[width=\linewidth]{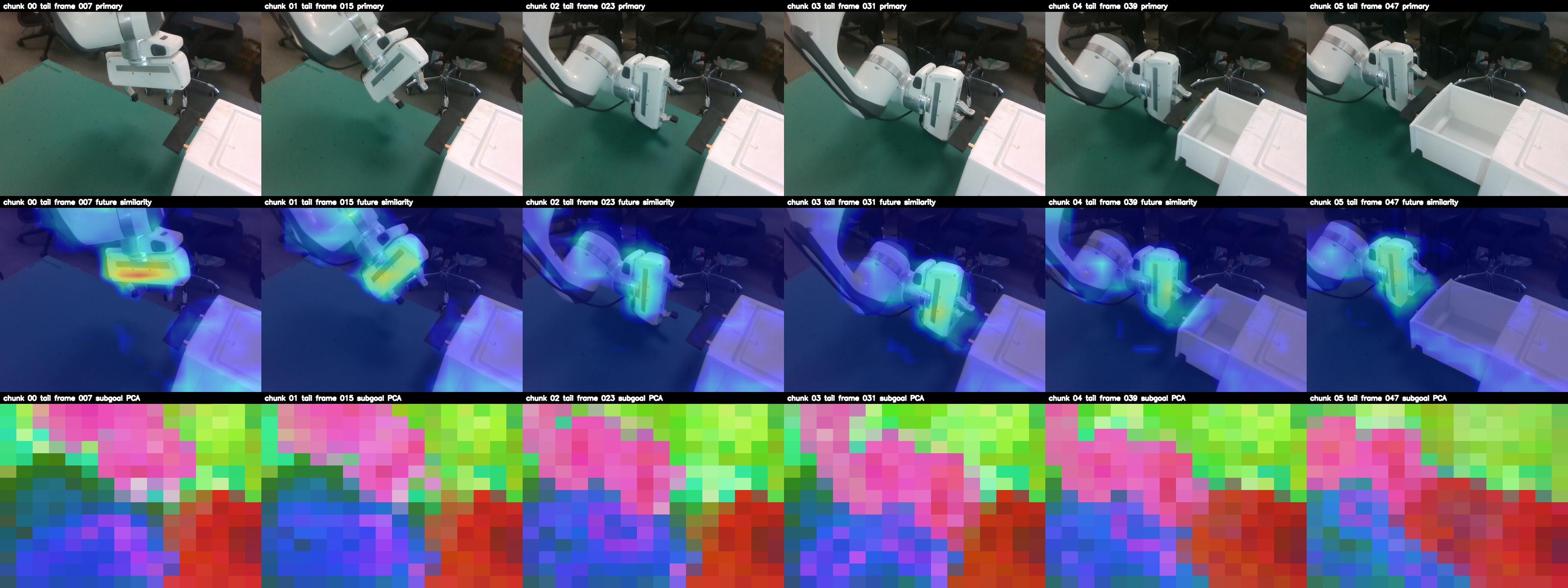}
	\caption{\textbf{LaWM subgoal visualization for Drawer Opening.} The Franka arm grasps the handle and pulls the drawer open.}
	\label{fig:opendrawer}
\end{figure}

\begin{figure}[t]
	\centering
	\includegraphics[width=\linewidth,height=0.88\textheight,keepaspectratio]{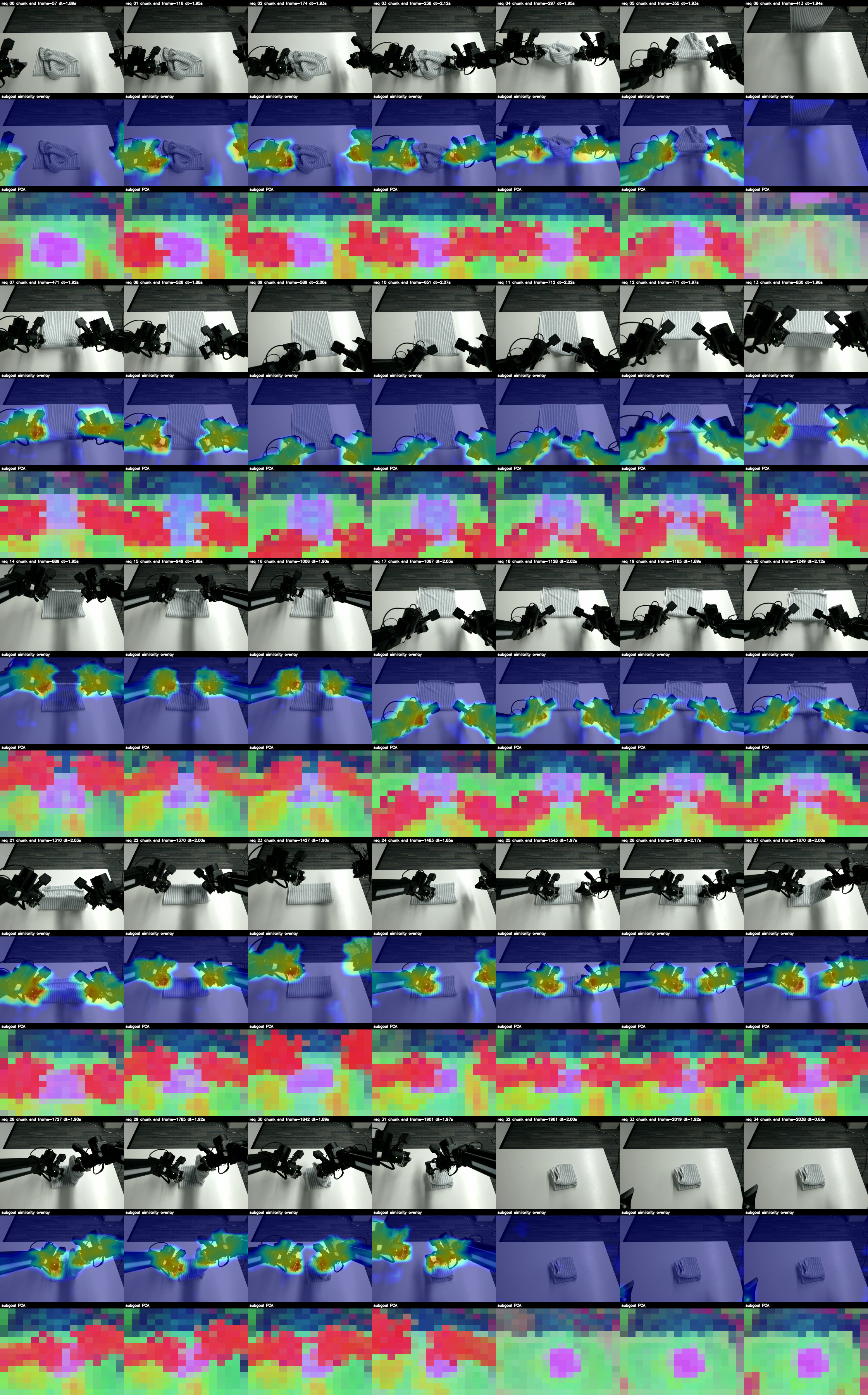}
	\caption{\textbf{LaWM subgoal visualization for Towel Folding.} Quanta X1 shakes out and flattens the towel, then completes two long-edge folds and one short-edge fold.}
	\label{fig:fold_towel_all}
\end{figure}

% \begin{table}[htbp]
%   \caption{Dataset ratios after merging selected subsets.}
%   \label{tab:dataset-ratios}
%   \centering
%   \begin{tabular}{clr}
%     \toprule
%     Idx & Dataset                         & Ratio (\%) \\
%     \midrule
%     0   & robocasa\_365\_pretrain         & 4.25       \\
%     1   & robocasa\_tabletop\_merged\_v30 & 4.39       \\
%     2   & BridgeV2                        & 5.26       \\
%     3   & fractal\_lerobot                & 5.53       \\
%     4   & droid\_1.0.1                    & 23.81      \\
%     5   & AgiBot\_merge                   & 12.44      \\
%     6   & robomind\_agilex\_3rgb          & 7.45       \\
%     7   & robomind\_franka\_3rgb          & 4.00       \\
%     8   & robomind\_ur\_1rgb              & 5.50       \\
%     9   & robomind\_franka\_fr3\_dual     & 0.55       \\
%     10  & epic\_kitchens\_100\_lerobot    & 2.40       \\
%     11  & EgoDex\_lerobot                 & 6.58       \\
%     12  & Egocentric-10K\_lerobot         & 17.86      \\
%     \bottomrule
%   \end{tabular}

% \end{table}

% \newpage

%===============================================================================

% \clearpage
% % The acknowledgments are automatically included only in the final and preprint versions of the paper.
% \acknowledgments{}

\end{document}